\documentclass[10pt,twocolumn,letterpaper]{article}

\usepackage[pagenumbers]{cvpr}      

\usepackage{graphicx}
\usepackage{amsmath}
\usepackage{amssymb}
\usepackage{booktabs}
\usepackage{booktabs}
\usepackage{multirow}
\usepackage{float}
\usepackage{amsthm,amsmath,amssymb}
\usepackage{mathrsfs}
\usepackage{bm}
\usepackage[pagebackref,breaklinks,colorlinks]{hyperref}
\usepackage[normalem]{ulem}
\useunder{\uline}{\ul}{}

\usepackage[table,xcdraw]{xcolor}

\usepackage[pagebackref,breaklinks,colorlinks]{hyperref}

\usepackage[capitalize]{cleveref}
\crefname{section}{Sec.}{Secs.}
\Crefname{section}{Section}{Sections}
\Crefname{table}{Table}{Tables}
\crefname{table}{Tab.}{Tabs.}

\def\cvprPaperID{2991} 
\def\confName{CVPR} 
\def\confYear{2023}

\begin{document}

\title{Detachable Novel Views Synthesis of Dynamic Scenes \\ Using Distribution-Driven Neural Radiance Fields}






\author{
Boyu Zhang$^{1}$, Wenbo Xu$^1$, Zheng Zhu$^{1}$, Guan Huang$^1$\\
$^1$PhiGent Robotics\\
{\tt\small \{wenbo.xu, guan.huang\}@phigent.ai, byz@seu.edu.cn, zhengzhu@ieee.org} 
}

\maketitle

\begin{abstract}
Representing and synthesizing novel views in real-world dynamic scenes from casual monocular videos is a long-standing problem. Existing solutions typically approach dynamic scenes by applying geometry techniques or utilizing temporal information between several adjacent frames without considering the underlying background distribution in the entire scene or the transmittance over the ray dimension, limiting their performance on static and occlusion areas. Our approach \textbf{D}istribution-\textbf{D}riven neural radiance fields offers high-quality view synthesis and a 3D solution to \textbf{D}etach the background from the entire \textbf{D}ynamic scene, which is called $\text{D}^4$NeRF. Specifically, it employs a neural representation to capture the scene distribution in the static background and a 6D-input NeRF to represent dynamic objects, respectively. Each ray sample is given an additional occlusion weight to indicate the transmittance lying in the static and dynamic components. We evaluate $\text{D}^4$NeRF on public dynamic scenes and our urban driving scenes acquired from an autonomous-driving dataset. Extensive experiments demonstrate that our approach outperforms previous methods in rendering texture details and motion areas while also producing a clean static background. Our code will be released \footnote{\href{https://github.com/Luciferbobo/D4NeRF}{https://github.com/Luciferbobo/D4NeRF}}.

\end{abstract}

\vspace{-0.4cm}
\section{Introduction}

Novel view synthesis \cite{avidan1997novel,avidan1998novel,criminisi2007efficient} from dynamic scenes aims to generate photorealistic views at arbitrary viewpoints and any time step with given images from one or more cameras as input. It provides the possibility to generate free-view rendering using finite-view input and brings a lifelike representation. This can have vast and varied applications, \eg switching views in cinematic/games special effects \cite{tewari2020state}, rendering visuals in AR/VR world \cite{2022vr1,2018vr2,2022Multiview}, assisting camera imaging \cite{2022Dark}, and enabling interactive exploration in robot/autonomous-driving perception and navigation \cite{2022Block,2022Panoptic}.

\setlength{\abovecaptionskip}{-0.15cm}
\begin{figure}[H]
\begin{center}
\includegraphics[width=0.485\textwidth]{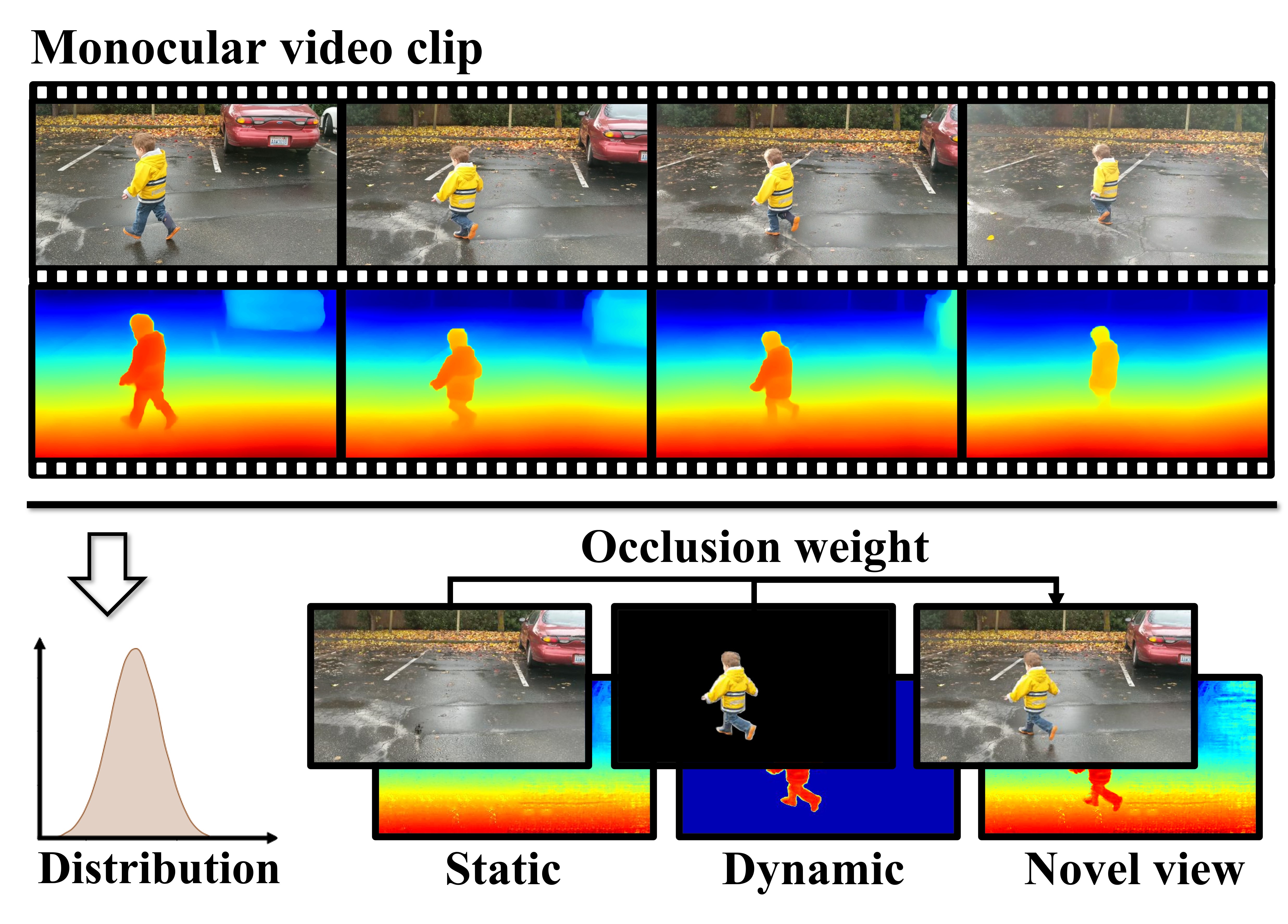} 
\end{center}
\caption{
Our method takes a casual monocular video clip captured from real-world dynamic scenes as input and generates a static representation driven by the underlying background and a dynamic representation. The target novel view rendering can be obtained by blending them with occlusion weight.}
\label{fig:appetizer}

\end{figure}

In essence, novel view synthesis of dynamic scenes is an extension of representing static 3D scenes from discrete 2D images, which is highly ill-posed \cite{ivanov2013theory} since there are infinite solutions that can render the input video appropriately while only a single 2D image observation is available at each view. Compared to static scenes \cite{2019DeepView,2017Single,2017Learning,2019Differentiable,2019Pushing,2018Stereo}, dynamic novel view synthesis is more challenging since approaches for dynamic scenes need to capture spatial-temporal information using a 6D plenoptic representation ($\mathbf{I}(t)=\Phi(\mathbf{x}, \mathbf{d},t ; \boldsymbol{\theta}): \mathbb{R}^3 \times \mathbb{R}^2\times \mathbb{R} \rightarrow \mathbb{R}^3$), and the occlusion between different objects might also be tricky.

Traditionally, novel view synthesis in real-world dynamic scenes \cite{2020Yoon,2020Blind,2020Consistent,2003Spacetime,oswald2014generalized,broxton2020immersive,collet2015high,orts2016holoportation} requires images from multiple camera views to estimate the geometry and occlusion of the entire scene. However, capturing such multi-view inputs is laborious and expensive. Neural Radiance Fields (NeRF) \cite{2020Nerf} offers a new circumvention to this problem when just given monocular views. Rather than explicit shape representation, NeRF encodes 3D location and viewing angle to color and density using a neural network and generates arbitrary novel views through volume rendering. Recent NeRF-based methods \cite{2021DNeRF,2021Gao,2021Nerfies,2021NRF4D,2021NSFF,2021NonRigid,2021Space,2021T,2022Panoptic,2022NeuMan,2022Multiview,2022Block,2022Fourier} show promising results. Using one or more NeRFs, these methods can implicitly represent dynamic scenes for novel view synthesis and time interpolation. 

Several methods \cite{2022Panoptic,2022NeuMan,2022Multiview,2022Block} use supervision networks (\eg semantic segmentation) to produce dynamic view synthesis and can achieve editable results, but they are limited to specific domains. Other methods \cite{2021Gao,2021Nerfies,2021NRF4D,2021NSFF,2021NonRigid} focus on adjacent temporal information in a finite number of frames to capture time-dependent components, such as optical flow and depth, while overlooking the underlying distribution existing in the entire video clip. For scenes containing a single object, such as Lego and Fern \cite{2020Nerf}, the 2D scene projection on the camera plane will alter drastically due to the switch of the view angle. Unlike these scenes, real-world locomotor scenes often contain a dynamic foreground and a static background. These static backgrounds, such as blocks, buildings, and other rigid areas, are far away from the camera lens. When photographing these scenes from various perspectives, the 2D projection on the camera plane will be only minimally disturbed by the change in viewpoint. It implies that the bulk of similar projection pixels in a real-world dynamic scene are shared by the static background across multiple frames. Assuming that identical regions are the observations for each frame and obey the scene distribution, then the camera projections from different angles can be obtained with tiny shifts based on this representation of distribution. Moreover, all previous approaches determine occlusion only from RGB space by merging rendering color and neglect transmittance weight existing on each ray sample, resulting in a sub-optimal performance in occlusion areas between the static backdrop and dynamic objects. 

To tackle the aforementioned distribution representation problem and provide occlusion weight over the ray dimension, we introduce $\text{D}^4$NeRF, a novel method that \textit{captures the underlying distribution in the entire scene and adds transmittance to each ray}. It can generate high-quality novel views at arbitrary viewpoints and any interpolation time steps for view synthesis of real-world dynamic scenes, and applied to general domains. Specifically, we extend NeRF to a parallel structure, as shown in \cref{fig:appetizer}. A background pipeline presenting the underlying distribution and a 6D-input NeRF generating time-varying fields are used to express static and dynamic components, respectively. Then we weight transmittance matrices on their rendering corresponding to each ray to learn the occlusion relationship. To optimize the pipeline effectively, we use multiple regularization losses to drive training in different modules. We evaluate $\text{D}^4$NeRF on NVIDIA dynamic scenes \cite{2020Yoon} and our urban driving scenes obtained from an autonomous-driving dataset, Argoverse\cite{Argoverse2}. Our contributions can be summarized as follows:

\begin{itemize}

\item To the best of our knowledge, we are the first to exploit the underlying scene distribution and produce transmittance across each ray sample in order to depict dynamic real-world scenes.

\item We present a novel attention-based structure and an occlusion weight neural representation that offers a 3D pattern for decoupling the clean static background from the entire scene. 

\item Substantial quantitative and qualitative experiments suggest that $\text{D}^4$NeRF operates better than existing methods. Additionally, an urban driving scenes dataset is built for dynamic novel view synthesis. We will release this dataset for research purpose.

\end{itemize}

\setlength{\belowcaptionskip}{-0.3cm} 
\begin{figure*}[t] 
\begin{center}
\includegraphics[width=0.95\textwidth]{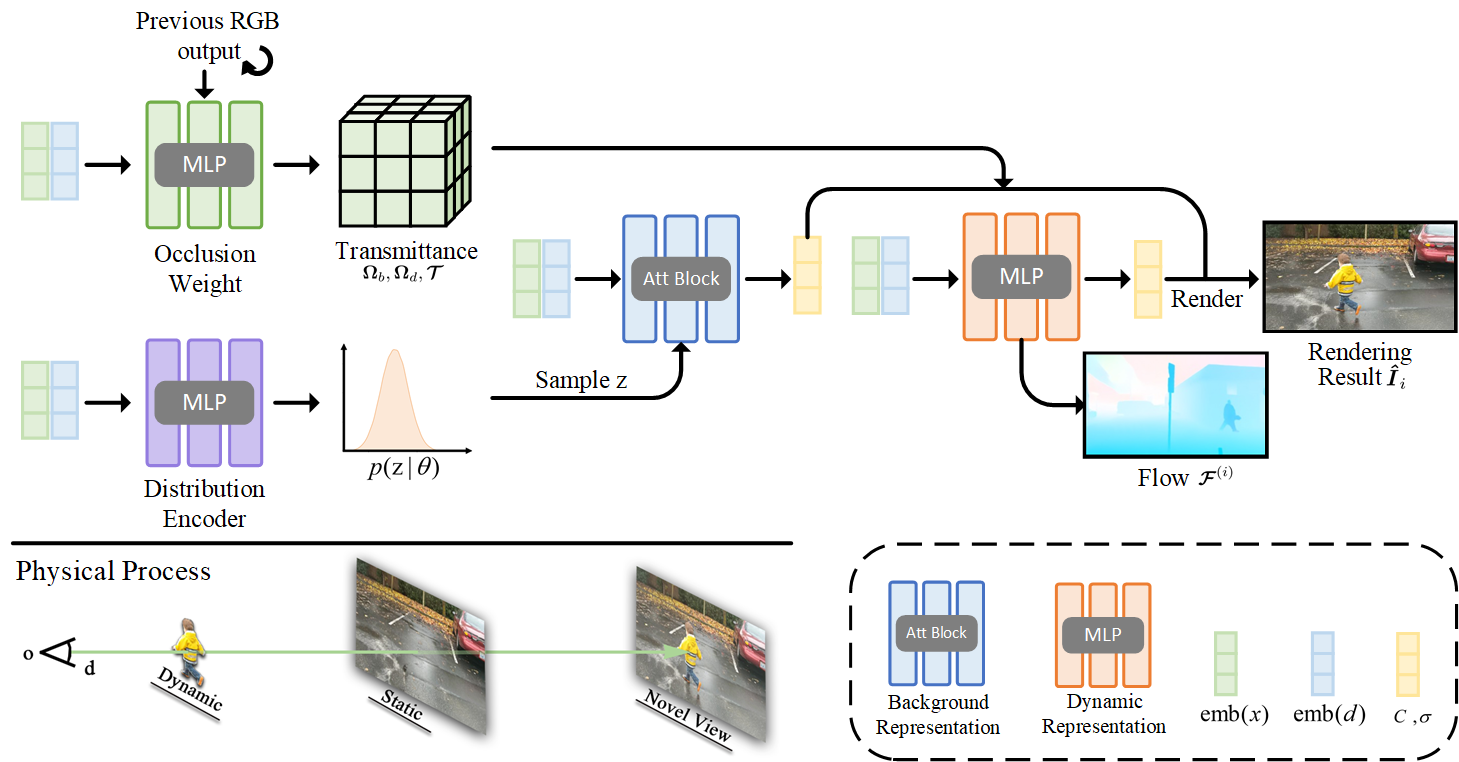} 
\end{center}
   \caption{\textbf{Network pipeline of $\text{D}^4$NeRF. } 
   Our approach trains two neural representations jointly: a distribution-driven {\color[HTML]{4672c4} background representation} and a flow-predicting {\color[HTML]{ed7d31} dynamic representation}, producing RGB color $C$ and volume density $\sigma$. A {\color[HTML]{937ce0} distribution encoder} is constructed to learn the implicit distribution from the background. And an {\color[HTML]{70ad47} occlusion weight module} is created to generate transmittance.
   At $i$-th time instance, the final rendering result $\bm{\mathit{\hat{I}}}_i$ is made by mixing occlusion weight $\Omega$ and $\bm{\mathcal{T}}$ with the outputs of two networks, as described in the physical process (bottom).}
\label{fig:main}
\vspace{-0.2cm}
\end{figure*}

\section{Related Work}

\noindent\textbf{Novel view synthesis.} 
The fundamental idea behind novel view synthesis is to generate the geometry of scenes from a set of limited views. Intuitively, we need to create an explicit 3D representation of scene geometry, such as point clouds or meshes \cite{buehler2001unstructured,chaurasia2013depth,debevec1996modeling,hedman2017casual,snavely2006photo}, and render novel views by transporting pixels among views via this representation. There are many works that concentrate on using various graphic techniques, such as light field \cite{chai2000plenoptic,gortler1996lumigraph,levoy1996light,penner2017soft,shi2014light,mildenhall2019local} and multi-plane based methods \cite{zhou2018stereo,choi2019extreme,riegler2020free,tucker2020single,srinivasan2019pushing}, to obtain 3D representation. In addition, learning-based methods \cite{chen2019learning,mescheder2019occupancy,michalkiewicz2019deep,park2019deepsdf,trevithick2020grf} attempt to utilize neural networks to learn view interpolation to obtain implicit representation. Meanwhile, neural networks can provide multiple priors to improve producing 3D representation, such as extracting meshes features \cite{riegler2020free,hedman2018deep} or estimating sparse/dense depth maps as prior \cite{flynn2018deepstereo,xu2019deep,niklaus20193d}. More recently, Neural Radiance Fields (NeRF) \cite{2020Nerf} have attained photorealistic synthesis performance by rendering novel views through MLP-based radiance and opacity fields. NeRF brings new enlightenment to novel view synthesis, and many methods try applying it to various scenes. NeRFw \cite{2021Nerfw} is suitable for novel view synthesis in outdoor scenes. Mip-NeRF \cite{barron2022mip} enhances NeRF for unbounded scenes. And many methods \cite{2022Urban,2022Mega,2022Bungee} extend NeRF to large-scale scenes. Nevertheless, these methods are only applicable to static scenes, as a 6D-input of dynamic scenes will lead to radiance ambiguity \cite{2021Nerf++}. 

\noindent\textbf{Video view synthesis for dynamic scenes.} Since the promising performance of neural radiance field, many works \cite{2021DNeRF,2021NSFF,2021NRF4D,2021Gao,yuan2021star,2021NonRigid,2021Nerfies} consider extending NeRF as a spatial-temporal representation to address view synthesis in dynamic scenes. Some of the previous methods \cite{kanade1997virtualized,zitnick2004high,li2012temporally} for dynamic view synthesis capture geometry and textures explicitly. Other previous learning-based models \cite{2020Yoon,huang2018deep,bansal20204d,bemana2020x} attempt to produce novel views using neural networks, while their performance is limited by the count of input views. Unlike these approaches, NeRF-based methods perform view synthesis for dynamic scenes with fewer input views and could enable time interpolation. D-NeRF \cite{2021DNeRF} considers time as an additional input to handle single dynamic objects indoors. \cite{2021NSFF,2021NRF4D} model the dynamic scene as a time-dependent continuous function with optical flow. Gao \etal \cite{2021Gao} introduce regularization losses to encourage plausible reconstruction. Xian \etal \cite{2021Space} employ video depth estimation to supervise a space-time radiance field. Using a ray-blending network, Tretschk \etal \cite{2021NonRigid} model the occlusion parts in dynamic scenes. TöRF \cite{2021T} utilizes prior information from the time-of-flight camera to enhance reconstruction quality. Park \etal \cite{2021Nerfies} optimize an additional continuous volumetric deformation field to capture dynamic humans. Li \etal \cite{2022Multiview} aim to represent scenes in multi-view videos effectively. These NeRF-based approaches take into account the use of temporal information between several successive frames, such as optical flow and depth, but disregard the global distribution over all frames. More recently, NeRF-based methods \cite{2022NeuMan,2022Panoptic} have provided a more accurate representation of dynamic scenes and achieved editable effects, but they are limited to specific domains.

\section{Method}

In this section, we introduce the proposed \textit{ \textbf{D}istribution-\textbf{D}riven Neural Radiance Fields for \textbf{D}etachable Novel Views Synthesis of \textbf{D}ynamic Scenes}, $\text{D}^4$NeRF. As shown in \cref{fig:main}, $\text{D}^4$NeRF consists of a background distribution driving module and a 6D-input NeRF. An occlusion weight module is designed to decouple the dynamic and static components. We first describe the background pipeline and the temporal-input NeRF in \cref{method_back} and \cref{method_dy}, respectively. Then the occlusion weight module will be discussed in \cref{method_w}. Lastly, a group of loss functions for optimizing the proposed network will be explained in \cref{method_loss}. 

\subsection{Preliminaries}
NeRF represents the radiance $\mathbf{c}=(r, g, b)$ and differential volume density $\sigma$ at a 3D location $\mathbf{x}=(x, y, z)$ of a scene observed from a viewing direction $\mathbf{d}=(\theta, \phi)$ as a continuous multi-variate function using a neural network, $\Phi:(\mathbf{x}, \mathbf{d}) \rightarrow$ $(\mathbf{c}, \sigma)$. To capture high-frequency signals \cite{tancik2020fourier} in inputs, NeRF uses a set of sine and cosine functions for increasing frequencies, $\gamma(\mathbf{x})=\left(\mathbf{x}, \cdots, \sin \left(2^k \pi \mathbf{x}\right), \cos \left(2^k \pi \mathbf{x}\right), \cdots\right), k \in\{0, \ldots, m-1\}$, where $m$ is a hyper-parameter that controls the total number of frequency bands. The pixel color $\hat{\mathbf{C}}$ can be rendered by integrating the radiance modulated by the volume density along the camera ray $\mathbf{r}(s)=\mathbf{o}+s \mathbf{d}$, with the origin $\mathbf{o}$ and direction $\mathbf{d}$ defined by the specified camera pose and intrinsic parameters:

\setlength{\abovedisplayskip}{4pt} 
\setlength{\belowdisplayskip}{4pt}

$$
\hat{\mathbf{C}}(\mathbf{r})=\int_{s_n}^{s_f} T(s) \sigma(\mathbf{r}(s)) \mathbf{c}(\mathbf{r}(s), \mathbf{d}) ds ,
$$
\begin{equation}
\text{where}\ \ T(s)=\exp \left(-\int_{s_n}^s \sigma(\mathbf{r}(p)) d p\right).
\label{eq:nerfin}
\end{equation}
$s_n$ and $s_f$ denote the bounds of the volume depth range, $T(s)$ corresponds to the accumulated transparency along that ray. The loss is then the difference between the reconstructed color $\hat{\mathbf{C}}$ and the ground truth color $\mathbf{C}$ corresponding to the pixel that ray $\mathbf{r}$ originated from:

\begin{equation}
\mathcal{L}=\sum_{\mathbf{r}}\|\hat{\mathbf{C}}(\mathbf{r})-\mathbf{C}(\mathbf{r})\|_2^2.
\vspace{-0.4cm}
\end{equation}

\setlength{\abovecaptionskip}{-0.1cm}

\begin{figure}[t]
\begin{center}
\includegraphics[width=0.46\textwidth]{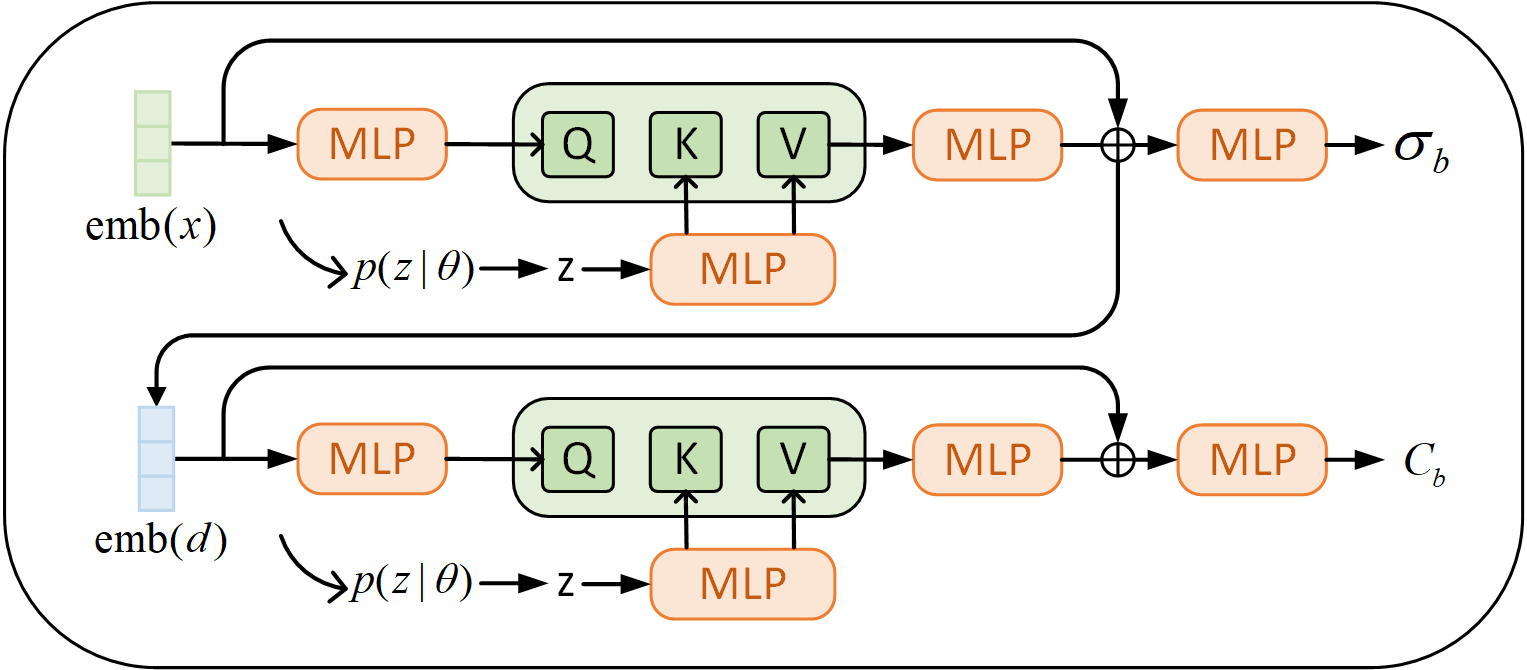} 
\end{center}
   \caption{\textbf{Attention block.} To compute attention, we use a latent variable sampled from the background distribution as \textbf{K}ey and \textbf{V}alue, and embedded 3D position and direction as \textbf{Q}uery. $\oplus$ denotes element-wise addition.
   }
\label{fig:back}
\end{figure}

\subsection{Distribution Representation}
\label{method_back}

We intend to represent the static background in terms of distribution such that camera projections from different views can be queried by small shifts based on this distribution representation. Assuming that the distribution obeys $P(\textbf{z}\in \mathbb{R}^D|\Theta)$, it can be formulated to:

\begin{equation}
\Phi_b:(\textbf{x},\textbf{d},\textbf{z})\rightarrow(\sigma_b,\textbf{c}_b).
\end{equation}

To obtain the implicit distribution from the static background, we construct a distribution encoder, as shown in \cref{fig:main}. Then a latent variable is sampled from the output posterior $\mathbf{z}\sim P_{\Theta}(\mathbf{z})$. The sample is utilized to drive the rendering of the background component. Due to the resultful performance of attention mechanisms\cite{vaswani2017attention, dosovitskiy2020image, liu2021swin}, we apply an attention-based structure to exploit the latent variable, as shown in \cref{fig:back}. 

Specifically, we apply attention to embedded inputs with querying sample $z$. By mapping embedded features and the latent distribution sample into a low-dimensional space, attention elements of 3D location $\mathbf{Q}_l, \mathbf{K}$,and $ \mathbf{V}$ can be derived (see supplemental material). $\textbf{F}_{\uparrow}$ denotes re-projecting inputs to the original-dimension space and extending elements along the ray dimension with interleaving repetition. Then the attention feature $\bm{\mathcal{C}}_l$ of 3D location can be interpreted as:

$$
\bm{\mathcal{C}}_l=\textbf{F}_{\uparrow}(\operatorname{Att}(\mathbf{Q}_l))+emb(\mathbf{x}),
$$
\begin{equation}
\text{where}\ \ \operatorname{Att}(\mathbf{Q}_l)=\operatorname{softmax}\left(\frac{\mathbf{Q}_l\mathbf{ K}^T}{\sqrt{d}}\right) \mathbf{V}.
\label{eq:att}
\end{equation}

Attention feature of view direction $\bm{\mathcal{C}}_d$ can be calculated in the same pattern as \cref{eq:att}, with $\bm{\mathcal{C}}_l$ concatenated to the input. Then the volumetric density and RGB color will be generated by subsequent networks $\mathbf{\sigma}_b=\textbf{MLP}(\bm{\mathcal{C}}_l)$ and $\mathbf{c}_b=\textbf{MLP}(\bm{\mathcal{C}}_d)$. Same to NeRF, the density part is solely determined by the position $\textbf{x}$. The rendered RGB color $\hat{\mathbf{C}}_b(\mathbf{r})$ is denoted as:

$$
\hat{\mathbf{C}}_b(\mathbf{r})=\int_{s_n}^{s_f} T(s) \sigma_b(\textbf{r}(s),\textbf{z}) \textbf{c}_b(\textbf{r}(s), \textbf{d}, \textbf{z})  d s
$$
\begin{equation}
\text{where}\ \ T(s)=\exp \left(-\int_{s_n}^s \sigma_b(\mathbf{r}(p),\textbf{z}) d p\right).
\end{equation}

\subsection{Motion Representation}
\label{method_dy}

We employ a dynamic NeRF to describe the time-varying motion in real-world dynamic scenes. At $i$-th time instance, we have:

\begin{equation}
\Phi_d:(\textbf{x},\textbf{d},i)\rightarrow(\sigma_d,\textbf{c}_d,\bm{\mathcal{F}}^{(i)})
\end{equation}

The rendered color $\hat{\mathbf{C}}^{i}(\mathbf{r})$ of the pixel corresponding to $i$-th time step is an integral over the radiance weighted by accumulated opacity $T(s)=\exp \left(-\int_{s_n}^s \sigma_d(\mathbf{r}(p),i) d p\right)$:

\begin{equation}
\hat{\mathbf{C}}^{(i)}_d(\mathbf{r})=\int_{s_n}^{s_f} T(s) \sigma_d\left(\mathbf{r}(s), i \right) \mathbf{c}_d\left(\mathbf{r}(s), \mathbf{d}, i \right) d s
\end{equation}

Dynamic NeRF also predicts scene flow in neighboring time instance to the current time step, $\bm{\mathcal{F}}^{(i)}=(\bm{f}^{(i)}_{\text{fw}},\bm{f}^{(i)}_{\text{bw}})$. Instead of rendering, this flow prediction is used to create region masks and geometric constraints, as explained in \cref{method_loss}.

At each time step, only one observation view is available. Consequently, generating the implicit representation of dynamic objects from a single video is an ill-posed problem. To effectively optimize our dynamic NeRF and maximize the usage of temporal information, we introduce multiple regularization losses described in \cref{method_loss}.

\subsection{Occlusion Weight}
\label{method_w}

Our background pipeline models the time-invariant component, and our dynamic NeRF generates a per-time-step deformation field. To combine them at pixel-level (ray-level), here we introduce our occlusion weight module, which targets rendering transmittance between static and dynamic components.

\begin{figure}[h]
\begin{center}
\includegraphics[width=0.40\textwidth]{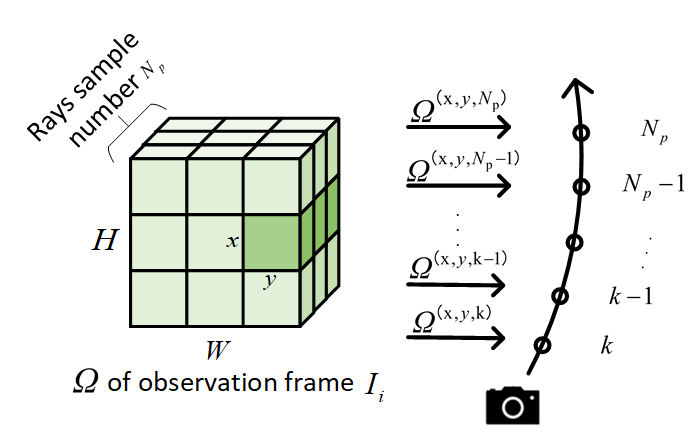} 
\end{center}
   \caption{\textbf{Transmittance on each ray.} 
   Each ray sample for a ray at observation time step $i$ at location $x,y$ is weighted by transmittance $\bm{\Omega}$ over the ray dimension.}
\label{fig:RW}
\end{figure}

\setlength{\abovecaptionskip}{0.2cm}

As shown in \cref{fig:RW}, at $i$-th frame, we calculate the transmittance $\bm{\Omega}_b,\bm{\Omega}_d \in \mathbb{R}^{H \times W \times N_p}$ corresponding to each sample for each ray and the transmittance $\bm{\mathcal{T}} \in \mathbb{R}^{H \times W}$  \footnote{In practice, limited by memory space, NeRF-based methods randomly choose $N_{\text{rand}}$ rays rather than rendering all $H\times W$ rays for an image at each training iteration.} corresponding to each ray using the warped previous rendering result $\bm{\mathit{I}}_{i-1}$ concatenated as input: 


\begin{equation}
\textbf{F}_w:(\textbf{x},\textbf{d},\bm{\mathit{I}}_{i-1})\rightarrow(\bm{\Omega}_b,\bm{\Omega}_d,\bm{\mathcal{T}}).
\end{equation}

The discrete integral formulation of rendered color $\hat{\mathbf{C}}_b(\mathbf{r})$ and  $\hat{\mathbf{C}}^{(i)}_d(\mathbf{r})$ are weighted along the ray dimension (see supplemental material), giving us the final rendered pixel at time step $i$ for location $x,y$:

\begin{equation}
\begin{aligned}
\hat{\mathbf{C}}^{(i)}(\mathbf{r})=& \bm{\mathcal{T}}^{(x,y)}\bm{\Omega}^{(x,y)}_b \hat{\mathbf{C}}_b(\mathbf{r})\\
&+(1-\bm{\mathcal{T}}^{(x,y)}) \bm{\Omega}^{(x,y)}_d \hat{\mathbf{C}}^{(i)}_d(\mathbf{r})
\end{aligned}
\end{equation}

Given the entire rays set $\mathcal{R} \in \mathbb{R}^{H \times W}$, the final rendering image can be calculated from  $\bm{\mathit{\hat{I}}}(x,y,i)=\{\hat{\mathbf{C}}^{(i)}(\mathbf{r})|\mathbf{r} \in \mathcal{R}\}$. Our experiments show that the occlusion weight helps decouple static and dynamic scene components.

\subsection{Loss}
\label{method_loss}

The lack of input views makes modeling time-dependent scenes ill-posed and difficult to optimize. To this end, several losses are designed to control pipeline training.

\noindent\textbf{Reconstruction loss.} 
 To learn the implicit representation from a $N$-frames input video clip  $\mathscr{D}=\left\{\left(\mathbf{x}^i, \mathbf{d}^i, \bm{\mathit{I}}^i\right)|i=1, \ldots,N\right\}$, we apply a reconstruction loss penalizing the difference between the target video frame $\mathbf{C}^{(i)}(\mathbf{r})$ and our yielding volume-rendered image $\hat{\mathbf{C}}^{(i)}(\mathbf{r})$:
 
\begin{equation}
\mathcal{L}_{\text {rec}}=\sum_{i=1}^{N}\sum_{\mathbf{r} \in \mathcal{R}}\left\|\hat{\mathbf{C}}^{(i)}(\mathbf{r})-\mathbf{C}^{(i)}(\mathbf{r})\right\|_2^2 
\end{equation}

\noindent\textbf{Distribution loss.}
Following Variational Auto-Encoder (VAE) \cite{kingma2013auto}, we impose scene distribution $P(\textbf{z}\in \mathbb{R}^D|\Theta)$ to a specified constant distribution. For Gaussian form, we take the mean component $\mu$ in VAE as our latent variable sample $P(\textbf{z}|\Theta)\sim \mathcal{N}(\mu,\sigma^2=0)$, and use the loss of this component throughout training:
 
\begin{equation}
\mathcal{L}_{\text {dist}}=\frac{1}{D}\sum_{i=1}^{D} -\frac{1}{2}(1-\mu^{2}_{i}).
\end{equation}


\noindent\textbf{Transmittance loss.}
Our occlusion weight module mitigates the ambiguous problem in occluded components between frames by rendering transmittance, targeting at predicting the overlay relationship for rendering results between the backdrop and dynamic foreground. Ideally, each ray should be rendered from either dynamic or static parts, hence the corresponding transmittance should be regularized to $0$ or $1$. Ray samples need to exist initially, and this encourages transmittance for each sample to be close to one:

\begin{equation}
\begin{aligned}
\mathcal{L}_{\text {w}}=&\sum_{\mathbf{r} \in \mathcal{R}}  -\bm{\mathcal{T}}^{(x,y)} \log (-\bm{\mathcal{T}}^{(x,y)}+\epsilon)\\
&+\sum_{\mathbf{r} \in \mathcal{R}} \sum_{k=1}^{N_p} \left\| 1-\bm{\omega}^{(k)}_{b}\right\|_1+\left\| 1-\bm{\omega}^{(k)}_{d}\right\|_1.
\end{aligned}
\end{equation}

Depth and optical flow serve as geometry supervision priors in many NeRF-based methods, aiding resolve motion–appearance ambiguity and accelerating convergence \cite{deng2022depth,2021Space,2021NSFF}. Here we use depth loss $\mathcal{L}_{\text{depth}}$ and flow consistence losses $\mathcal{L}_{\text{cons}}, \mathcal{L}_{\text{flow}}$ for regularization, which will be discussed detailedly in supplemental material. With using $\lambda$ coefficients weight each term, the overall loss can be interpreted as:
\begin{equation}
\begin{aligned}
\mathcal{L}_{\text {overall }}=&\lambda_{\text {\text{rec} }}\mathcal{L}_{\text{rec}}+\lambda_{\text {\text{dist} }}\mathcal{L}_{\text{dist}}
+\lambda_{\text{depth}}\mathcal{L}_{\text{depth}}\\ 
&+\lambda_{\text{cons}} \mathcal{L}_{\text{cons}}
+\lambda_{\text{flow}} \mathcal{L}_{\text{flow}}+\lambda_{\text{w}} \mathcal{L}_{\text{w}}
\end{aligned}
\vspace{-0.2cm}
\end{equation}

\setlength{\belowcaptionskip}{-0.15cm} 

\begin{figure*}
\begin{subfigure}{0.42\linewidth}
    \includegraphics[width=\textwidth]{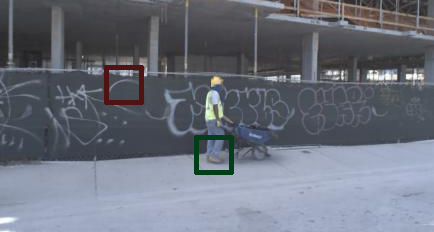}
\end{subfigure}
\begin{subfigure}{0.111\linewidth}
    \includegraphics[width=\textwidth]{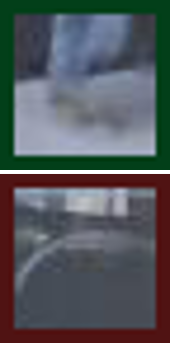}
\end{subfigure}
\begin{subfigure}{0.111\linewidth}
    \includegraphics[width=\textwidth]{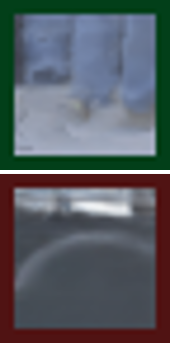}
\end{subfigure}
\begin{subfigure}{0.111\linewidth}
    \includegraphics[width=\textwidth]{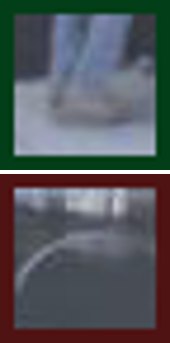}
\end{subfigure}
\begin{subfigure}{0.111\linewidth}
    \includegraphics[width=\textwidth]{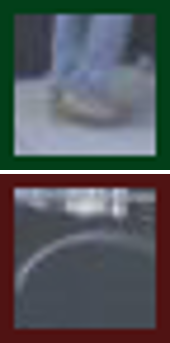}
\end{subfigure}
\begin{subfigure}{0.111\linewidth}
    \includegraphics[width=\textwidth]{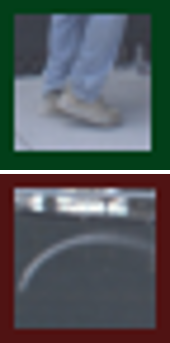}
\end{subfigure}

\begin{subfigure}{0.42\linewidth}
    \includegraphics[width=\textwidth]{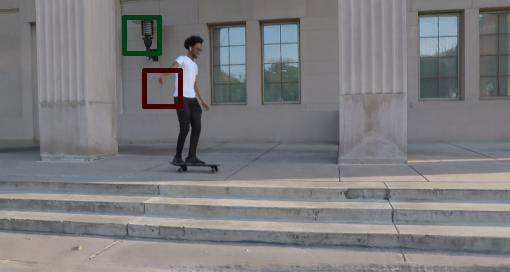}
\end{subfigure}
\begin{subfigure}{0.111\linewidth}
    \includegraphics[width=\textwidth]{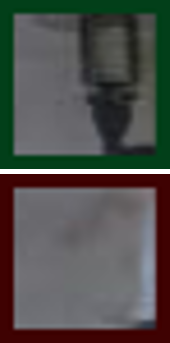}
\end{subfigure}
\begin{subfigure}{0.111\linewidth}
    \includegraphics[width=\textwidth]{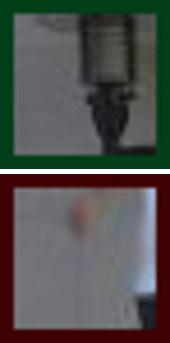}
\end{subfigure}
\begin{subfigure}{0.111\linewidth}
    \includegraphics[width=\textwidth]{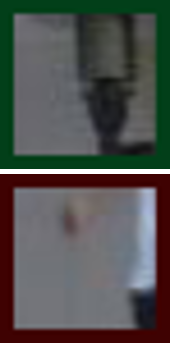}
\end{subfigure}
\begin{subfigure}{0.111\linewidth}
    \includegraphics[width=\textwidth]{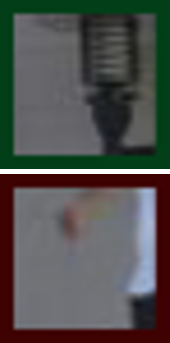}
\end{subfigure}
\begin{subfigure}{0.111\linewidth}
    \includegraphics[width=\textwidth]{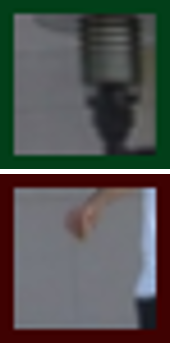}
\end{subfigure}

\begin{subfigure}{0.42\linewidth}
    \includegraphics[width=\textwidth]{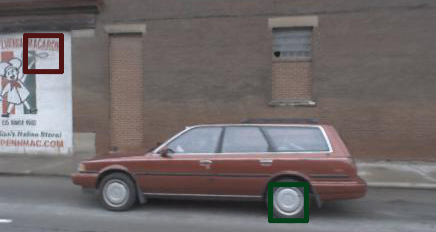}
    \caption{Ours}
\end{subfigure}
\begin{subfigure}{0.111\linewidth}
    \includegraphics[width=\textwidth]{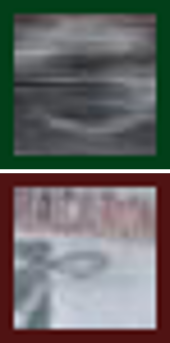}
    \caption{NeRF}
\end{subfigure}
\begin{subfigure}{0.111\linewidth}
    \includegraphics[width=\textwidth]{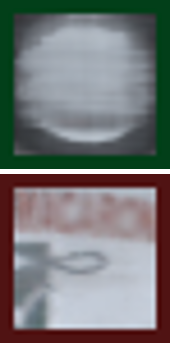}
    \caption{Gao \etal}
    \label{fig:com1gao}
\end{subfigure}
\begin{subfigure}{0.111\linewidth}
    \includegraphics[width=\textwidth]{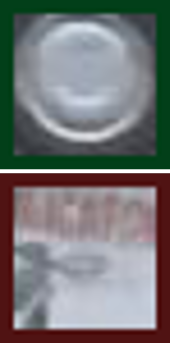}
    \caption{NSFF}
\end{subfigure}
\begin{subfigure}{0.111\linewidth}
    \includegraphics[width=\textwidth]{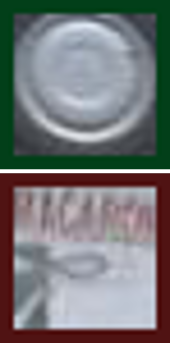}
    \caption{Ours}
\end{subfigure}
\begin{subfigure}{0.111\linewidth}
    \includegraphics[width=\textwidth]{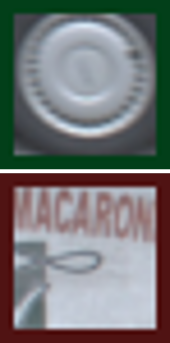}
    \caption{Reference}
\end{subfigure}

\caption{\textbf{Comparison of novel view synthesis details.} Our method shows a more effective way of recovering texture in both static and motion regions.}
\label{fig:com1}
\vspace{-0.25cm}
\end{figure*}

\begin{figure*}
\begin{subfigure}{0.247\linewidth}
    \includegraphics[width=\textwidth]{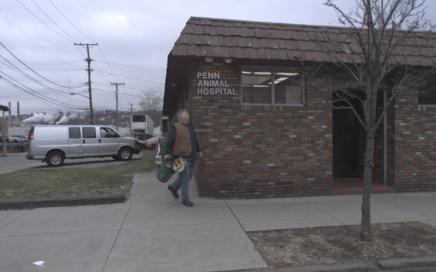}
\end{subfigure}
\begin{subfigure}{0.247\linewidth}
    \includegraphics[width=\textwidth]{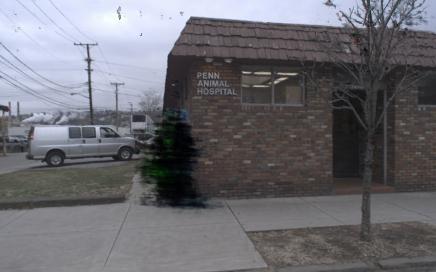}
\end{subfigure}
\begin{subfigure}{0.247\linewidth}
    \includegraphics[width=\textwidth]{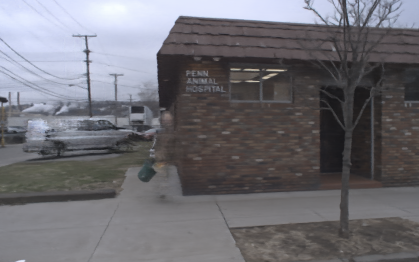}
\end{subfigure}
\begin{subfigure}{0.247\linewidth}
    \includegraphics[width=\textwidth]{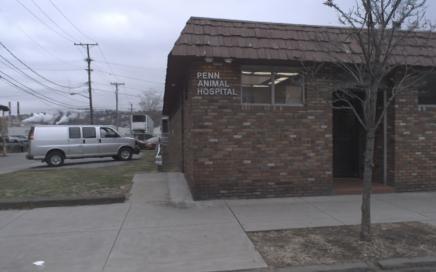}
\end{subfigure}

\begin{subfigure}{0.247\linewidth}
    \includegraphics[width=\textwidth]{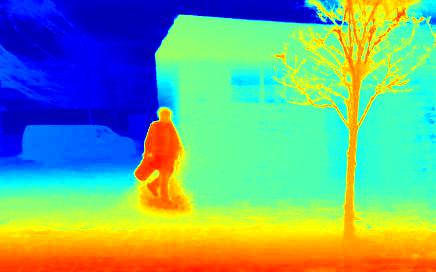}
\end{subfigure}
\begin{subfigure}{0.247\linewidth}
    \includegraphics[width=\textwidth]{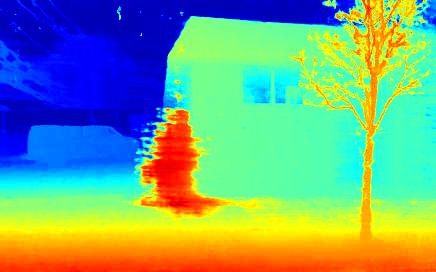}
\end{subfigure}
\begin{subfigure}{0.247\linewidth}
    \includegraphics[width=\textwidth]{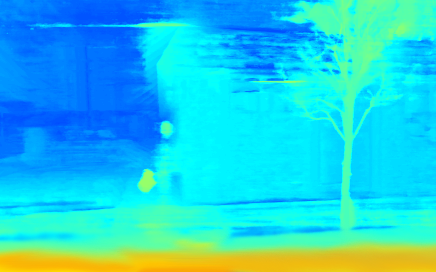}
\end{subfigure}
\begin{subfigure}{0.247\linewidth}
    \includegraphics[width=\textwidth]{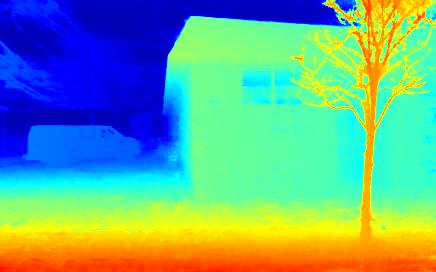}
\end{subfigure}

\begin{subfigure}{0.247\linewidth}
    \includegraphics[width=\textwidth]{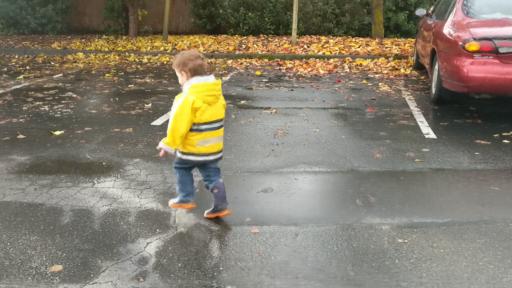}
\end{subfigure}
\begin{subfigure}{0.247\linewidth}
    \includegraphics[width=\textwidth]{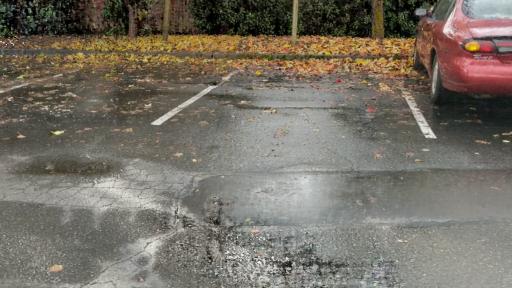}
\end{subfigure}
\begin{subfigure}{0.247\linewidth}
    \includegraphics[width=\textwidth]{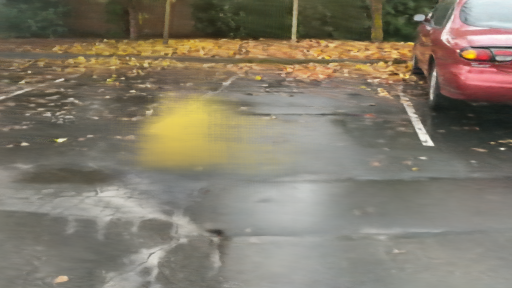}
\end{subfigure}
\begin{subfigure}{0.247\linewidth}
    \includegraphics[width=\textwidth]{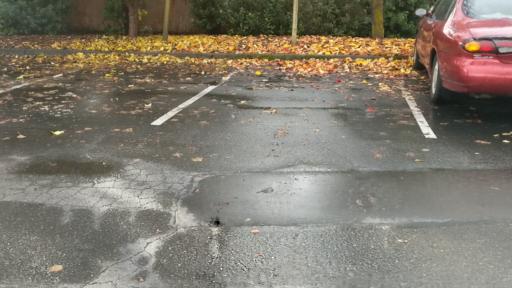}
\end{subfigure}

\begin{subfigure}{0.247\linewidth}
    \includegraphics[width=\textwidth]{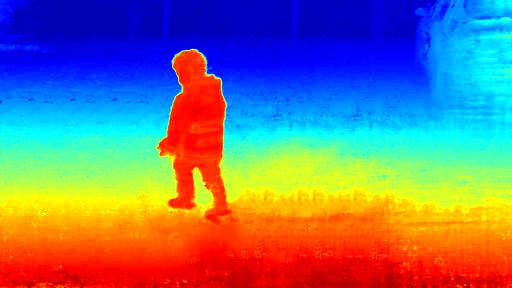}
\end{subfigure}
\begin{subfigure}{0.247\linewidth}
    \includegraphics[width=\textwidth]{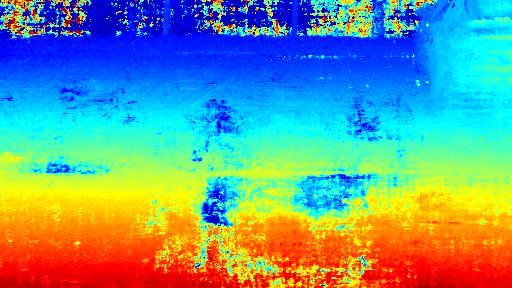}
\end{subfigure}
\begin{subfigure}{0.247\linewidth}
    \includegraphics[width=\textwidth]{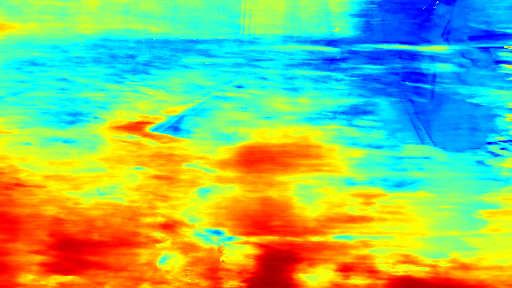}
\end{subfigure}
\begin{subfigure}{0.247\linewidth}
    \includegraphics[width=\textwidth]{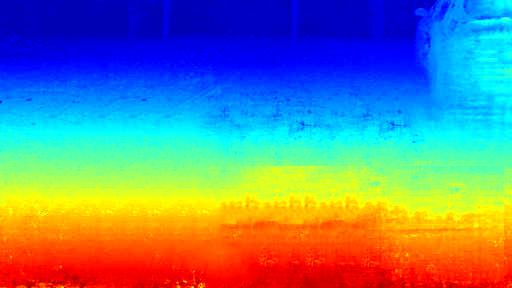}
\end{subfigure}

\begin{subfigure}{0.247\linewidth}
    \includegraphics[width=\textwidth]{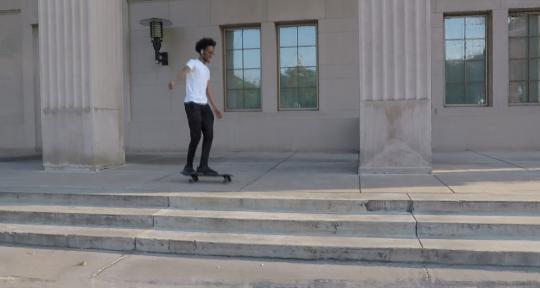}
\end{subfigure}
\begin{subfigure}{0.247\linewidth}
    \includegraphics[width=\textwidth]{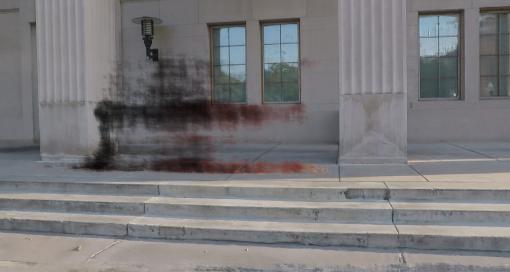}
\end{subfigure}
\begin{subfigure}{0.247\linewidth}
    \includegraphics[width=\textwidth]{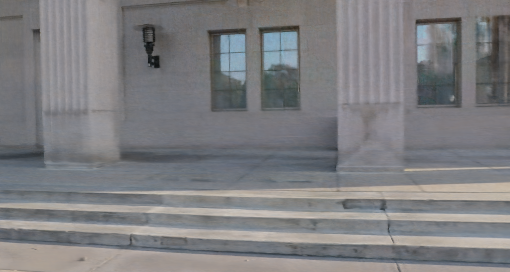}
\end{subfigure}
\begin{subfigure}{0.247\linewidth}
    \includegraphics[width=\textwidth]{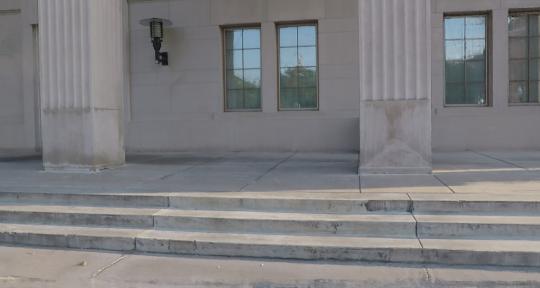}
\end{subfigure}

\begin{subfigure}{0.247\linewidth}
    \includegraphics[width=\textwidth]{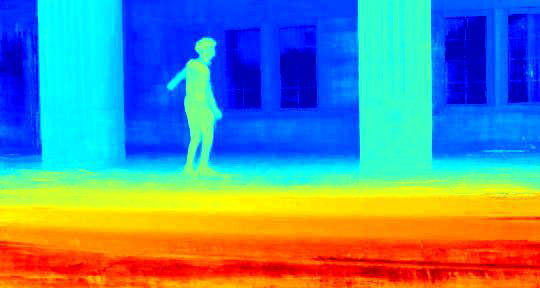}
    \caption{Entire Scene}
\end{subfigure}
\begin{subfigure}{0.247\linewidth}
    \includegraphics[width=\textwidth]{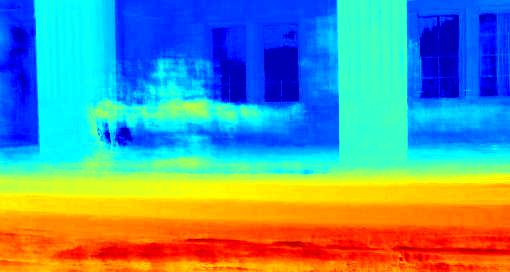}
    \caption{NSFF}
\end{subfigure}
\begin{subfigure}{0.247\linewidth}
    \includegraphics[width=\textwidth]{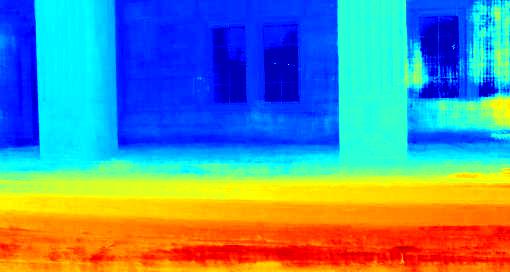}
    \caption{Gao \etal}
\end{subfigure}
\begin{subfigure}{0.247\linewidth}
    \includegraphics[width=\textwidth]{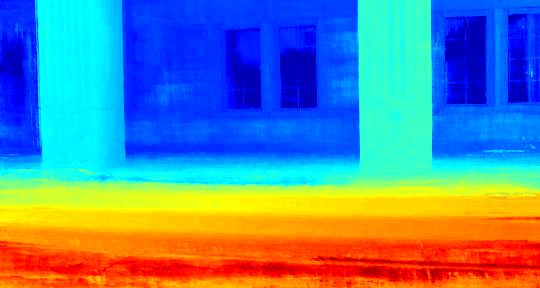}
    \caption{Ours}
    \label{fig:decoupled}
\end{subfigure}
\caption{\textbf{Comparison on decoupling static background from entire scenes.} (a) shows the entire scene, (b-d) shows the separation results of the static background from NSFF, Gao \etal, and ours. $\text{D}^4$NeRF obtains an uncontaminated background in both RGB and depth space.}
\label{fig:decouple}
\vspace{-0.15cm}
\end{figure*}

\section{Experiments}

\subsection{Datasets and Metrics}

\noindent\textbf{Urban driving scenes.} We built a dataset of real-world street scenes captured from an autonomous-driving collection Argoverse \cite{Argoverse2}. All scenes are taken in urban street, using one of seven high-resolution surrounded monocular cameras recorded at 30 Hz. Each gathered video clip depicts an outdoor scene with vehicles or people performing dynamic actions, spans 20-50 frames , and is scaled to a resolution of $436 \times 272$. We use corresponding estimation methods (See \cref{imp}) to acquire depth, optical flow, and camera poses. Since monocular cameras cannot provide multiple views, we carry out time interpolation for evaluation in each scene. The recording frequency is decreased to 15 Hz for training.

\noindent\textbf{NVIDIA dynamic scenes.} We then evaluate the dynamic scenes from NVIDIA \cite{2020Yoon}, which comprise 4 scenes captured by a hand-held monocular moving camera and 8 scenes captured by 12 stationary multi-view cameras evenly distributed and manually synchronized. 7 multi-view scenes are selected as our evaluation target for their correct camera pose estimation. For each scene, 20 to 30 frames are extracted from the source video for training. At each time instance, just one of these cameras is utilized as input, and other 11 camera views per time step are used for evaluation. Selected clips are downsized to a resolution of $510 \times 272$.

\noindent\textbf{Metrics.} All comparison approaches are evaluated by PSNR, SSIM \cite{wang2004image} and LPIPS \cite{zhang2018perceptual}. Higher PSNR and SSIM imply superior, while lower LPIPS indicate better.

\subsection{Implementation Details}
\label{imp}

All experiments are carried out on 8 NVIDIA 3090 GPUs in the Pytorch framework. We use 48 threads of Intel(R) Xeon(R) Platinum 8255C CPU @ 2.50GHz to expedite data loading. The weight coefficients for $\mathcal{L}_{\text{rec}},\ \mathcal{L}_{\text{dist}},\  \mathcal{L}_{\text{depth}},\ \mathcal{L}_{\text{cons}},\  \mathcal{L}_{\text{flow}}$, and $\mathcal{L}_{\text {w }}$ are 1, 5e-1, 3e-2, 3e-2, 1e-2, and 5e-1, respectively. Adam optimizer \cite{kingma2014adam} with $\beta_1=0.9, \beta_2=0.999$, and $\epsilon$=1e-8 is used with learning rate of 3e-4. We generate $N_{\text{rand}}=1024$ rays randomly in a batch for each training iteration, and we sample $N_p=128$ ray points along each ray. We initially warm up our occlusion weight module to get the right flow estimation for 2e5 iterations supervised by ground truth flow estimated from RAFT \cite{teed2020raft}. Thereafter, all losses are involved to train another 5e5 iterations. To get ground truth depth, we leverage the state-of-the-art monocular image depth estimation method \cite{ranftl2020towards}. COLMAP SfM technique \cite{schonberger2016structure} is applied to estimate the camera poses, scene bounds, and camera params with the premise that intrinsics and extrinsics are fixed.

\subsection{Quantitative Evaluation}

To quantitatively validate the performance of our method, we carry out experiments on NVIDIA dynamic scenes and urban driving scenes. We compare $\text{D}^4$NeRF with four baselines. The first is a classical learning-based method \cite{2020Yoon}. The second is vanilla NeRF \cite{2020Nerf}. The third and the fourth are recent state-of-the-art methods \cite{2021NSFF,2021Gao} that also use multiple NeRFs to perform dynamic novel view synthesis. All baseline methods are implemented using the author’s provided official codebase and default hyper-parameters.

\cref{table:compare1} shows the performance of our method on NVIDIA dynamic scenes compared to other baseline methods. $\text{D}^4$NeRF achieves the best performance on average and in most scenes. Due to space limitations, we only present the results of PSNR and LPIPS on NVIDIA dynamic scenes, and please refer to our supplemental material for complete comparison results of all three metrics on both datasets.

\begin{table*}[t]
\setlength{\tabcolsep}{0.3mm}{
\begin{tabular}{@{}ccccccccccccccccc@{}}
\toprule
                   & \multicolumn{2}{c}{Skating}                                   & \multicolumn{2}{c}{Balloon1}                                  & \multicolumn{2}{c}{Balloon2}                                  & \multicolumn{2}{c}{Truck}                                     & \multicolumn{2}{c}{Umbrella}                                  & \multicolumn{2}{c}{Jumping}                                   & \multicolumn{2}{c}{Playground}                                & \multicolumn{2}{c}{Ave}                                        \\ \cmidrule(l){2-17} 
\multirow{-2}{*}{} & PSNR                          & LPIPS                         & PSNR                          & LPIPS                         & PSNR                          & LPIPS                         & PSNR                          & LPIPS                         & PSNR                          & LPIPS                         & PSNR                          & LPIPS                         & PSNR                          & LPIPS                         & PSNR                          & LPIPS                          \\ \midrule
Yoon \etal.\cite{2020Yoon}               & 24.33                             & .1563                             & 19.33                             & .1625   & 19.78                             & .1797                             & 28.78                             & .0835                            & 20.37                              & .1696                          & 21.51                             & .2294                            & 17.43                             & .2124                             & 21.65                             & .1705                              \\
NeRF\cite{2020Nerf}               & 25.40                         & .0906                         & 21.41                         & .1412                         & 23.30                         & .0705                         & 27.93                         & .0975                         & 21.23                         & .2341                         & 22.21                         & .1511                         & 20.75                         & .1566                         & 23.18                         & .1345                          \\
NSFF\cite{2021NSFF}               & 33.42                         & .0235                         & 23.39                         & .0762                         & 27.85                         & .0490                         & \textbf{32.09} & .0372                         & 23.67                         & .1153                         & 26.67                         & .0624                         & 23.77                         & .0857                         & 27.26                         & .0642                         \\
Gao \etal \cite{2021Gao}                & {\ul 33.83} & {\ul .0225} & {\ul 23.60} & {\ul .0736} & \textbf{27.90} & {\ul .0457} & 31.00                         & {\ul.0321} & \textbf{24.25} & {\ul.1071} & {\ul 26.72} & {\ul .0584} & {\ul 23.82} & {\ul .0823} & {\ul 27.30} & {\ul .0603} \\
Ours               & \textbf{34.52} & \textbf{.0171} & \textbf{23.87} & \textbf{.0666} & {\ul27.60} & \textbf{.0411} & {\ul31.75} & \textbf{.0232} & {\ul24.20} & \textbf{.1037} & \textbf{27.00} & \textbf{.0463} & \textbf{23.94} & \textbf{.0734} & \textbf{27.55} & \textbf{.0530} \\ \bottomrule
\end{tabular}}
\caption{\textbf{Quantitative  comparison results on NVIDIA dynamic scenes.} The \textbf{best} result is in bold, and the {\ul second-best} is underlined in each column.}
\label{table:compare1}
\vspace{-1.0em}
\end{table*}

\begin{table}[t]
\setlength{\tabcolsep}{1.3mm}{
\begin{tabular}{@{}l|c|c|c|c|c@{}}
\toprule
Method      & DT & RW & $\mathcal{L}_{\text{flow}}$ & $\mathcal{L}_{\text{depth}}$ & PSNR/LPIPS \\ \midrule
Base        &              &            &   &   &     25.23/.1144        \\
Base+DT   & \checkmark            &            &   &   &  26.02/.0899          \\
Base+DT+RW & \checkmark            & \checkmark          &   &   & { 27.44/.0655 }           \\
Ours+$\mathcal{L}_{\text{flow}}$      & \checkmark            & \checkmark          & \checkmark &   &{\ul 27.50/.0601 }           \\
Ours+$\mathcal{L}_{\text{flow}}$+$\mathcal{L}_{\text{depth}}$    & \checkmark           & \checkmark          & \checkmark & \checkmark &\textbf{27.55/.0530 }          \\ \bottomrule
\end{tabular}}
\caption{\textbf{Ablation study.} We evaluate different modules on NVIDIA dynamic scenes dataset. PSNR and LPIPS are shown on average.}
\label{table:ablation}
\end{table}

\noindent\textbf{Ablation study.} We verify the effectiveness of different modules in our approach, including the distribution encoder (DT), the ray occlusion weight module (RW), and two regularization losses $\mathcal{L}_{\text{depth}}$ and $\mathcal{L}_{\text{flow}}$. We conduct our ablation study on NVIDIA dynamic scenes shown in \cref{table:ablation}. The first row shows the performance of only two naive NeRFs, while the second row proves the effectiveness of DT, which brings $0.79\uparrow$ and $.0245\downarrow$ improvement on PSNR and LPIPS. Besides, RW also shows an increase of $5.5\%$ on PSNR and a decrease of $27.1\%$ on LPIPS. The depth loss and the flow loss further show their efficacy in the fourth and fifth rows, respectively. 
Our ablation study shows that the model performs best when all modules are used.

\begin{figure}[t]
\begin{subfigure}{0.32\linewidth}
    \includegraphics[width=\textwidth]{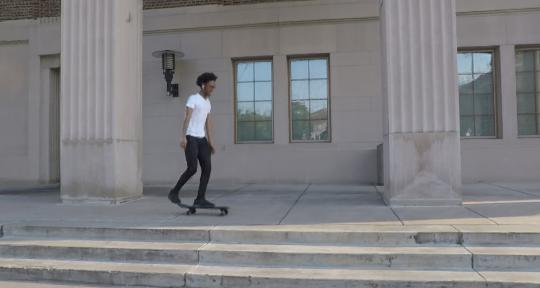}
    \caption{Entire Scene}
\end{subfigure}
\begin{subfigure}{0.32\linewidth}
    \includegraphics[width=\textwidth]{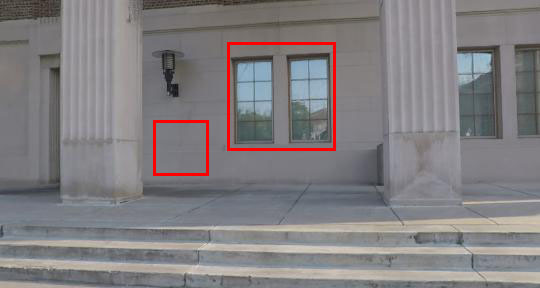}
    \caption{Ours}
\end{subfigure}
\begin{subfigure}{0.32\linewidth}
    \includegraphics[width=\textwidth]{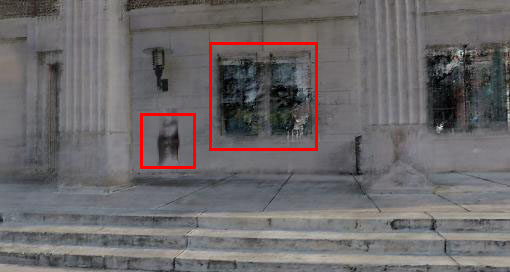}
    \caption{w/o RW}
\end{subfigure}
\caption{\textbf{Visual ablation on the occlusion weight module.} RW ensures proper separating capacity.}
\label{fig:decouple_ab}
\vspace{-0.25cm}
\end{figure}

\subsection{Qualitative Evaluation}

While this section has many qualitative evaluations, we strongly advise the reader to watch the supplementary video for a more thorough look.

In \cref{fig:com1}, we compare novel view rendering visually. $\text{D}^4$NeRF surpasses all other methods on dynamic scenes. NeRF is incapable of handling dynamic regions and causes significant ghosting. NSFF recovers dynamic objects properly but produces imprecise details, such as the wall in the first scene and the lamp in the second scene. Gao \etal capture finer details but does poorly in recovering exact motion. Our method presents the highest qualitative results in recovering static texture and motion details.

\noindent\textbf{Background decoupling.} We further show that our approach could decouple static backgrounds from entire dynamic scenes self-supervised. $\text{D}^4$NeRF can restore the unoccluded clean background simply by observing the occluded regions in different frames of a video clip. Other NeRF-based methods also tend to infer occlusion relationships, but get subpar performance. As seen in \cref{fig:decouple}, compared to other multi-NeRF-based methods, $\text{D}^4$NeRF obtains the most accurate separation results in RGB-D space with fewer artifacts and blurs. In \cref{fig:decouple_ab}, we also show ablation that without our occlusion weight, the decoupled synthesized results are fuzzy and noisy. 

Though the separation of static components from dynamic scenes has been addressed in 2D geometric algorithms, the geometry and technique of inpainting lack 3D comprehension, leading to restricted results in arbitrary view synthesis and time interpolation. Our approach gives a 3D pattern for dealing with this problem and allows the reconstruction of a decoupled implicit 3D scene representation. This makes it possible to generate novel views at arbitrary viewpoints and any input time step.

\setlength{\belowcaptionskip}{-0.1cm} 

\begin{figure}[t]
\begin{subfigure}{0.455\linewidth}
    \includegraphics[width=\textwidth]{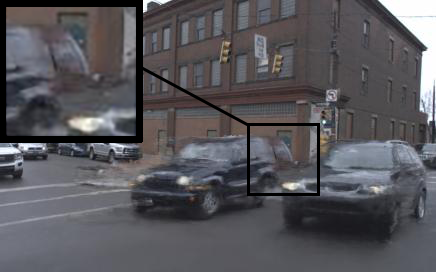}
    \caption{Ghosting in motion areas.}
\end{subfigure}
\begin{subfigure}{0.535\linewidth}
    \includegraphics[width=\textwidth]{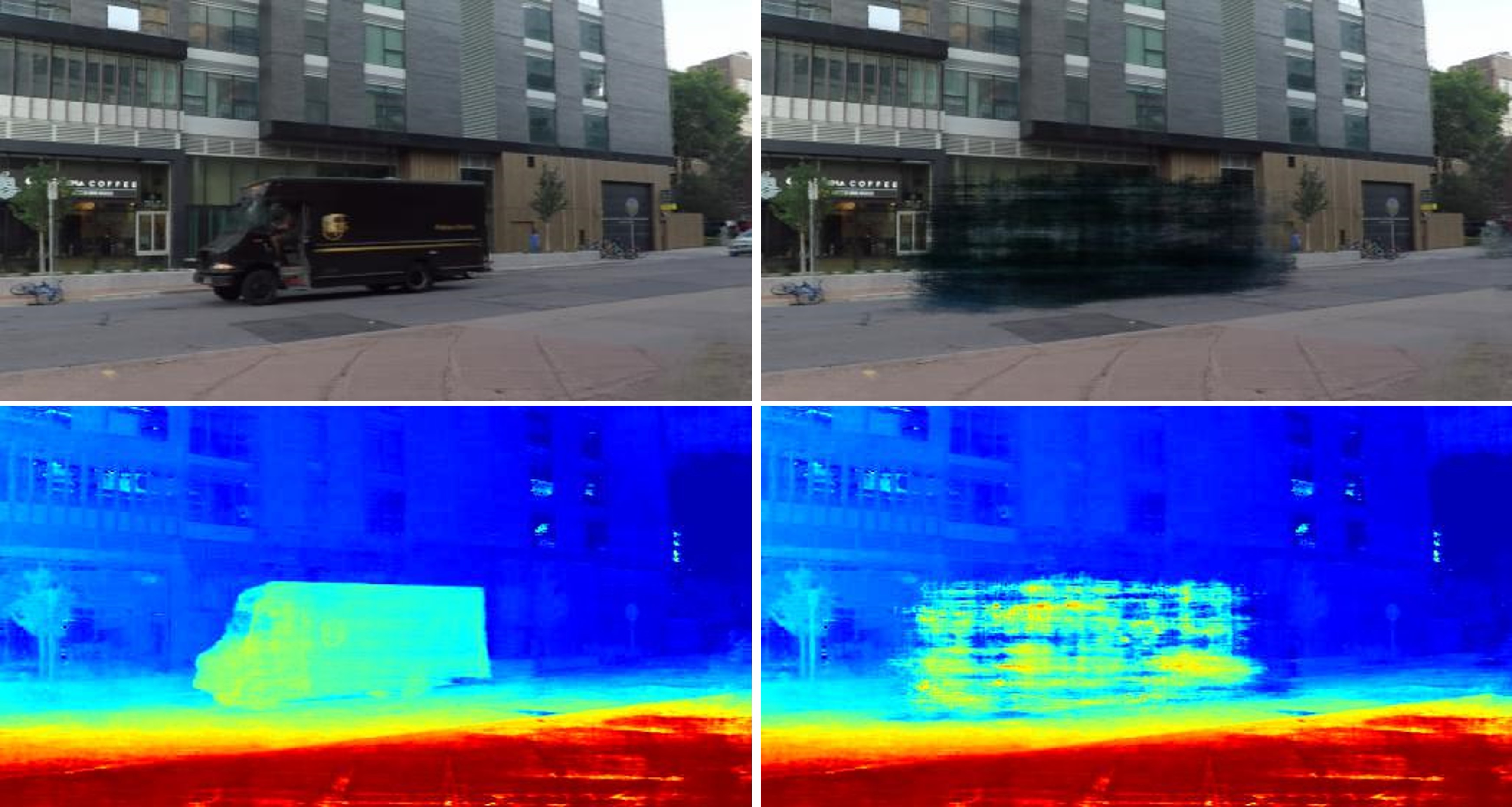}
    \caption{Half-baked decouple results.}
\end{subfigure}
\caption{\textbf{Limitations.} Our method fails to recover the motion area (a) and is unable to produce a complete decoupling result (b) with incorrect flow or depth estimation.}
\label{fig:failure}
\vspace{-0.25cm}
\end{figure}

\section{Conclusion}

We presented $\text{D}^4$NeRF, a method for modeling dynamic scenes in the real world and performing novel view synthesis from
casual monocular videos. $\text{D}^4$NeRF provides a 3D pattern that could decouple the occlusion area into static and dynamic components and recovers a clean background. We empirically demonstrate superior quantitative and qualitative performance on both urban driving scenes and NVIDIA dynamic scenes. Our work suggests various directions for future research, such as exploring high-quality decomposition and editor models for 3D dynamic scenes. In addition, we notice that some approaches \cite{2021Nerfies,park2021hypernerf} use deformable latent code to control the representation of dynamic regions. The same can be applied in $\text{D}^4$NeRF to generate delicate deformations for dynamic objects.

\noindent\textbf{Limitations.} Same as other NeRF-based methods, our approach relies on accurate estimation, \eg camera poses and optical flow, to enable the proper representation of dynamic and static parts. Inaccurate ones may lead to motion artifacts and incorrect decoupling results (See \cref{fig:failure}).

\appendix

\section{Details of Rendering Formulations}

\subsection{Attention Block}

\begin{figure}[h]
\begin{center}
\includegraphics[width=0.45\textwidth]{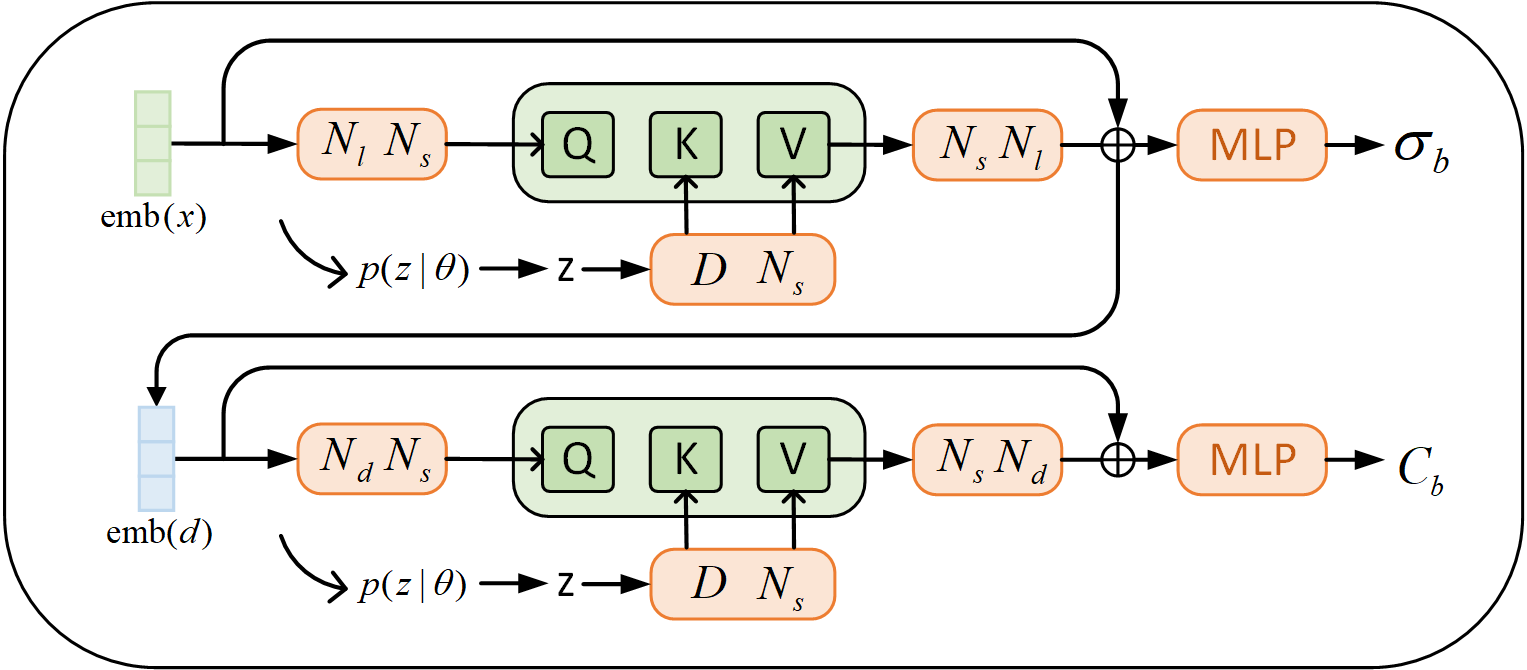} 
\end{center}
   \caption{\textbf{Detailed Pipeline of Attention block.} The numbers under each network layer represent the input and output channels. $\oplus$ denotes element-wise addition.
   }
\label{fig:back}
\end{figure}

Here we describe attention block detailedly. To numerically estimate the continuous rendering integral, we adopt the same sampling technique and embedding function in NeRF\cite{2020Nerf}. The volume depth range can be partitioned into $N_p$ evenly-spaced bins, and we draw one sample uniformly and randomly from within each bin. Input embedding 3D location and view direction are written as $emb(x)\in \mathbb{R}^{N_{l}\times N_p}, emb(d)\in \mathbb{R}^{N_{d}\times N_p}$. $N_l$ and $N_d$ are the dimensions extended by embedding. We use fully connected layers $\textbf{F}_{\downarrow}:\mathbb{R}^{* \times N_p}\rightarrow \mathbb{R}^{N_{s} \times N_p}$ to project inputs into a lower-dimension space $N_{s}$.
In order to reduce calculation overhead, both in terms of time and memory space, we accumulate the latent features along with rays via average pooling over the ray dimension $\textbf{F}_p(\textbf{x})=\frac{1}{N} \sum_{i=1}^{N} \textbf{x}[:, i]:\mathbb{R}^{* \times N_p}\rightarrow \mathbb{R}^{*}$. The attention can be expressed as:

\begin{equation}
\begin{aligned}
\textbf{Q}_l &=\mathbf{F}_p(\mathbf{F}_{\downarrow}(emb(\mathbf{x}))), \\
\textbf{Q}_d &=\mathbf{F}_p(\mathbf{F}_{\downarrow}(emb(\mathbf{d}))), \\
\textbf{K} &=\textbf{V}=\mathbf{F}_p(\mathbf{F}_{\downarrow}(\mathbf{z})).
\end{aligned}
\end{equation}

We use $\textbf{F}_{\uparrow}:\mathbb{R}^{N_{s}}\rightarrow \mathbb{R}^{* \times N_p}$ to re-project inputs into the original-dimension space and extend elements along the ray dimension with interleaving repetition. The attention feature of location is denoted as:

$$
\bm{\mathcal{C}}_l=\textbf{F}_{\uparrow}(\operatorname{Att}(\mathbf{Q}_l))+emb(\mathbf{x}),
$$
\begin{equation}
\text{where}\ \ \operatorname{Att}(\mathbf{Q}_l)=\operatorname{softmax}\left(\frac{\mathbf{Q}_l\mathbf{ K}^T}{\sqrt{d}}\right) \mathbf{V}.
\label{eq:att}
\end{equation}

\subsection{Discrete Integral with Occlusion Weight}

The discrete integral formulations of rendered color $\hat{\mathbf{C}}_b(\mathbf{r})$, $\hat{\mathbf{C}}^{(i)}_d(\mathbf{r})$ are weighted by $\bm{\omega}_{b}$ and $\bm{\omega}_{d}$ over the ray dimension:

$$
\bm{\Omega}^{(x,y)}_b \hat{\mathbf{C}}_b(\mathbf{r})=\sum_{k=1}^{N_p}\left(T_k\left(1-e^{-\sigma^{(k)} \delta^{(k)}}\right) \mathbf{c}^{(k)}\bm{\omega}^{(k)}_{b}\right),
\label{eq:discrte_b}
$$

$$
\bm{\Omega}^{(x,y)}_d \hat{\mathbf{C}}^{(i)}_d(\mathbf{r}) =\sum_{k=1}^{N_p}\left(T_k\left(1-e^{-\sigma^{(k)} \delta^{(k)}}\right) \mathbf{c}^{(i,k)}\bm{\omega}^{(k)}_{b}\right),
\label{eq:discrte_d}
$$

\begin{equation}
\text{where}\ \ T_k=e^{-\sum_{j=1}^{k-1} \sigma^{(j)} \delta^{(j)}}, \delta_i=s_{i+1}-s_i. 
\end{equation}

\begin{table*}[!ht]
\setlength{\tabcolsep}{2.1mm}{
\begin{tabular}{@{}ccccccccccccc@{}}
\toprule
                   & \multicolumn{3}{c}{Skating}                                                                   & \multicolumn{3}{c}{Balloon1}                                                                  & \multicolumn{3}{c}{Balloon2}                                                                   & \multicolumn{3}{c}{Truck}                                                                     \\ \cmidrule(l){2-13} 
\multirow{-2}{*}{} & PSNR                          & SSIM                          & LPIPS                         & PSNR                          & SSIM                          & LPIPS                         & PSNR                          & SSIM                          & LPIPS                         & PSNR                          & SSIM                          & LPIPS                         \\ \midrule
Yoon \etal.\cite{2020Yoon}              & 24.33                             & .8273                             & .1563                             & 19.33                             & .7324                            & .1625                             & 19.78                             & .7458                             & .1797                             & 28.78                             & .9451                             & .0835                             \\
NeRF \cite{2020Nerf}              & 25.40                         & .9419                         & .0906                         & 21.41                         & .8630                         & .1412                         & 23.30                         & .9198                         & .0705                         & 27.93                         & .9289                         & .0975                         \\
NSFF \cite{2021NSFF}              & 33.42                         & .9754                         & .0235                         & 23.39                         & {\ul.8822} & .0762                         & {\ul 27.85} & \textbf{.9355} & .0490                         & \textbf{32.09} & .9618                         & .0372                         \\
Gao  \etal.\cite{2021Gao}              & {\ul33.83} & {\ul.9782} & {\ul.0225} & {\ul23.60} & .8795                         & {\ul.0736} & \textbf{27.90} & .9342                         & {\ul.0457} & 31.00                         & {\ul.9628} & {\ul.0321} \\
Ours               & \textbf{34.52} & \textbf{.9824} & \textbf{.0171} & \textbf{23.87} & \textbf{.8835} & \textbf{.0666} & 27.60                         & {\ul.9348} & \textbf{.0411} & {\ul31.75} & \textbf{.9652} & \textbf{.0232} \\ \bottomrule
\end{tabular}}

\caption{\textbf{Quantitative  comparison results on Skating, Balloon1, Balloon1 and Truck.} The \textbf{best} result is in bold, and the {\ul second-best} is underlined in each column.}
\label{table:c1}
\end{table*}

\begin{table*}[!ht]
\setlength{\tabcolsep}{2.1mm}{
\begin{tabular}{@{}ccccccccccccc@{}}
\toprule
                   & \multicolumn{3}{c}{Umbrella}                                                                    & \multicolumn{3}{c}{Jumping}                                                                     & \multicolumn{3}{c}{Playground}                                                                  & \multicolumn{3}{c}{Ave}                                                                         \\ \cmidrule(l){2-13} 
\multirow{-2}{*}{} & PSNR                          & SSIM                           & LPIPS                          & PSNR                          & SSIM                           & LPIPS                          & PSNR                          & SSIM                           & LPIPS                          & PSNR                          & SSIM                           & LPIPS                          \\ \midrule
Yoon  \etal.\cite{2020Yoon}             & 20.37                            & .7344                              & .1696                              & 21.51                             & .7928                              & .2294                             & 17.43                             & .5479                             & .2124                              & 21.65                             & .7608                              & .1705                              \\
NeRF  \cite{2020Nerf}             & 21.23                         & .7793                         & .2341                         & 22.21                         & .8466                         & .1511                         & 20.75                         & .8253                         & .1566                         & 23.18                         & .8721                         & .1345                         \\
NSFF  \cite{2021NSFF}             & 23.67                         & .8320                         & .1153                         & 26.67                         & {\ul.9220} & .0624                         & 23.77                         & .8707                         & .0857                         & 27.26                         & .9114                         & .0642                         \\
Gao   \etal.\cite{2021Gao}              & \textbf{24.25} & \textbf{.8445} & {\ul.1071} & {\ul26.72} & .9132                         & {\ul.0584} & {\ul23.82} & {\ul.8766} & {\ul.0823} & {\ul27.30} & {\ul.9127} & {\ul.0602} \\
Ours               & {\ul24.20} & {\ul.8323} & \textbf{.1037} & \textbf{27.00} & \textbf{.9262} & \textbf{.0463} & \textbf{23.94} & \textbf{.8795} & \textbf{.0734} & \textbf{27.55} & \textbf{.9148} & \textbf{.0531} \\ \bottomrule
\end{tabular}}

\caption{\textbf{Quantitative  comparison results on Umbrella, Jumping and playground.} Last column shows the average performance on NVIDIA dynamic scenes. The \textbf{best} result is in bold, and the {\ul second-best} is underlined in each column.}
\label{table:c2}
\end{table*}

\begin{table*}[!ht]
\centering
\setlength{\tabcolsep}{3.5mm}{
\begin{tabular}{@{}cccccccccc@{}}
\toprule
                                                        & \multicolumn{3}{c}{S1}                            & \multicolumn{3}{c}{S2}                           & \multicolumn{3}{c}{S3}                           \\ \cmidrule(l){2-10} 
                                                        & PSNR           & SSIM           & LPIPS           & PSNR           & SSIM           & LPIPS          & PSNR           & SSIM           & LPIPS          \\ \midrule
Yoon  \etal \cite{2020Yoon} & 35.33          & .9542          & .0544          & 25.31          & .8955          & .1526          & 28.77          & .9014          & .1138          \\
NeRF  \cite{2020Nerf}                     & 36.40          & .9720          & .0349          & 24.90          & .8846          & .1384          & 30.87          & .9152          & .0871          \\
NSFF  \cite{2021NSFF}                     & 39.56          & {\ul .9902}    & {\ul .0094}    & 31.41          & .9525          & .0398          & \textbf{32.19} & \textbf{.9441} & \textbf{.0529} \\
Gao   \etal.\cite{2021Gao}  & {\ul 40.61}    & .9897          & .0112          & {\ul 31.56}    & {\ul .9544}    & {\ul .0393}    & {\ul 31.55}    & .9365          & .0636          \\
Ours                                                        & \textbf{40.92} & \textbf{.9921} & \textbf{.0081} & \textbf{31.91} & \textbf{.9594} & \textbf{.0367} & 31.45          & {\ul .9372}    & {\ul .0537}    \\ \bottomrule
\end{tabular}}
\caption{\textbf{Quantitative  comparison results on S1, S2 and S3.} The \textbf{best} result is in bold, and the {\ul second-best} is underlined in each column.}
\label{table:u1}
\end{table*}

\begin{table*}[!ht]
\centering
\setlength{\tabcolsep}{3.5mm}{
\begin{tabular}{@{}cccccccccc@{}}
\toprule
                                                            & \multicolumn{3}{c}{S4}                           & \multicolumn{3}{c}{S5}                           & \multicolumn{3}{c}{Ave}                           \\ \cmidrule(l){2-10} 
                                                            & PSNR           & SSIM           & LPIPS          & PSNR           & SSIM           & LPIPS          & PSNR            & SSIM           & LPIPS          \\ \midrule
Yoon  \etal \cite{2020Yoon} & 25.12          & .8829          & .1409          & 31.40          & .9651          & .0562          & 29.19          & .9198          & .1036          \\
NeRF  \cite{2020Nerf}                      & 25.26          & .8848          & .1323          & 33.66          & .9736          & .0366          & 30.22          & .9260          & .0859          \\
NSFF  \cite{2021NSFF}                      & 26.20          & .9176          & .0790          & {\ul 34.67}    & {\ul .9803}    & {\ul .0254}    & 32.81          & .9569          & {\ul .0413}          \\
Gao   \etal.\cite{2021Gao}  & \textbf{26.53} & \textbf{.9259} & \textbf{.0608} & 34.02          & .9788          & .0327          & {\ul 32.85}    & {\ul .9571}    & { .0415}    \\
Ours                                                        & {\ul 26.40}    & {\ul .9225}    & {\ul .0616}    & \textbf{35.10} & \textbf{.9818} & \textbf{.0196} & \textbf{33.16} & \textbf{.9586} & \textbf{.0359} \\ \bottomrule
\end{tabular}}
\caption{\textbf{Quantitative  comparison results on S4, S5.} Last column shows the average performance on urban street scenes. The \textbf{best} result is in bold, and the {\ul second-best} is underlined in each column.}
\label{table:u2}
\end{table*}

\section{Regularization Loss}

\noindent\textbf{Depth loss.}
Depth serves as a geometry prior in many NeRF-based methods, and helps resolve motion–appearance ambiguity and accelerates convergence \cite{deng2022depth,2021Space}. We use depth loss to supervise the geometry learning of the entire scene. For neural rendering, depth is generated as an additional output using $\hat{\mathbf{D}}^{(i)}(\mathbf{r})=\int_{s_n}^{s_f} T(s) \sigma(\mathbf{r}(s), i)s\ d s$. Our depth loss aims at minimizing the difference between rendered depth $\hat{\mathbf{D}}^{(i)}(\mathbf{r})$ and ground truth depth $\mathbf{D}^{(i)}(\mathbf{r})$:

\begin{equation}
\mathcal{L}_{\text {depth }}=\sum_{i=1}^{N}\sum_{\mathbf{r} \in \mathcal{R}}\left\|{\hat{\mathbf{D}}^{(i)}(\mathbf{r})}-{\mathbf{D}^{(i)}(\mathbf{r})}\right\|_2^2
\end{equation}

\noindent\textbf{Optical flow loss.}
Optical flow presents a general method to describe dynamic motion. At $i$-th time step, for a pixel located at $(x,y,z)$, flow is defined as $\bm{f}=\left(f_x, f_y, f_z\right)=\left(\frac{\partial x}{\partial t}, \frac{\partial y}{\partial t}, \frac{\partial z}{\partial t}\right)$. Different from the semantic segmentation models for a specific single domain \cite{2022NeuMan,2022Panoptic}, flow-based segmentation provides a generalized paradigm for decoupling motion and static. We generate masks of dynamic objects and static backgrounds using optical flow estimation:

$$
\hat{\mathcal{M}}^{(i)}(\mathbf{r})= \begin{cases}1, & \text { dynamic } \\ 
0, & \text { static  }\end{cases}
$$

Warping manipulates the current frame for distorting position to adjacent frames. With optical flow, the future position of a continuous point $\bm{v}^{(i-1)}\in \mathbb{R}^3$ at timestamp $i-1$ is obtained through:

\begin{equation}
\begin{aligned}
\hat{\bm{v}}^{(i)}&=warp(\bm{v}^{(i-1)},i)=\bm{v}^{(i-1)}+\bm{f}^{(i-1)}_{\text{fw}}. 
\end{aligned}
\end{equation}

We use the operator $warp(\cdot)$ to denote warping. To maintain temporal photometric consistency in neighboring time steps, we force the rendering results of adjacent frames to be consistent, constraining forward and backward simultaneously:

\begin{equation}
\begin{aligned}
\mathcal{L}_{\text {cons}}=&\sum_{i=2}^{N}\sum_{\mathbf{r} \in \mathcal{R}}   \hat{\mathcal{M}}^{(i)}(\mathbf{r})\left\|warp\left(\hat{\mathbf{C}}^{(i-1)}(\mathbf{r}),i\right)-\mathbf{C}^{(i)}(\mathbf{r})\right\|_2^2 \\
&+ \sum_{i=1}^{N-2}\sum_{\mathbf{r} \in \mathcal{R}} \hat{\mathcal{M}}^{(i)}(\mathbf{r})\left\|warp\left(\hat{\mathbf{C}}^{(i+1)}(\mathbf{r}),i\right)-\mathbf{C}^{(i)}(\mathbf{r})\right\|_2^2
\end{aligned}
\end{equation}

Generating optical flow correctly from scratch is also challenging. Similar to methods \cite{2021NRF4D,2021NSFF,2021Gao} regularizing geometric meaning on optical flow, we minimize the difference between forward and backward flow and the penalty term on flow:

\begin{equation}
\begin{aligned}
\mathcal{L}_{\text {flow}}=&\sum_{i=1}^{N-1}\sum_{\mathbf{r} \in \mathcal{R}}   \left\|  \bm{f}^{(i)}_{\text{fw}} - \bm{f}^{(i+1)}_{\text{bw}}   \right\|_2^2\\
&+\frac{1}{N-1}\sum_{i=2}^{N-1}\sum_{\mathbf{r} \in \mathcal{R}} \left(  \left\|  \bm{f}^{(i)}_{\text{fw}}\right\|_1 + \left\|  \bm{f}^{(i)}_{\text{bw}}\right\|_1 \right).
\end{aligned}
\label{eq:flow}
\end{equation}

\section{Additional Comparison Results}

Here we offer more comparison results on NVIDIA dynamic scenes dataset and urban street scenes dataset, evaluating with all three comparison metrics PSNR, SSIM, and LPIPS. Comparisons for each scene on NVIDIA dataset are shown in \cref{table:c1} and \cref{table:c2}. Evaluation results in urban street scenes are presented in \cref{table:u1} and \cref{table:u2}.

{\small
\bibliographystyle{ieee_fullname}
\bibliography{egbib}

\begin{thebibliography}{10}\itemsep=-1pt

\bibitem{2021T}
Benjamin Attal, Eliot Laidlaw, Aaron Gokaslan, Changil Kim, and Matthew
  O'Toole.
\newblock T"orf: Time-of-flight radiance fields for dynamic scene view
  synthesis.
\newblock In {\em Advances in Neural Information Processing Systems (NeurIPS)},
  2021.

\bibitem{avidan1997novel}
Shai Avidan and Amnon Shashua.
\newblock Novel view synthesis in tensor space.
\newblock In {\em Proceedings of the IEEE Conference on Computer Vision and
  Pattern Recognition (CVPR)}, 1997.

\bibitem{avidan1998novel}
Shai Avidan and Amnon Shashua.
\newblock Novel view synthesis by cascading trilinear tensors.
\newblock {\em IEEE Transactions on Visualization and Computer Graphics
  (TVCG)}, 4(4):293--306, 1998.

\bibitem{bansal20204d}
Aayush Bansal, Minh Vo, Yaser Sheikh, Deva Ramanan, and Srinivasa Narasimhan.
\newblock 4d visualization of dynamic events from unconstrained multi-view
  videos.
\newblock In {\em Proceedings of the IEEE Conference on Computer Vision and
  Pattern Recognition (CVPR)}, 2020.

\bibitem{barron2022mip}
Jonathan~T Barron, Ben Mildenhall, Dor Verbin, Pratul~P Srinivasan, and Peter
  Hedman.
\newblock Mip-nerf 360: Unbounded anti-aliased neural radiance fields.
\newblock In {\em Proceedings of the IEEE Conference on Computer Vision and
  Pattern Recognition (CVPR)}, 2022.

\bibitem{bemana2020x}
Mojtaba Bemana, Karol Myszkowski, Hans-Peter Seidel, and Tobias Ritschel.
\newblock X-fields: Implicit neural view-, light-and time-image interpolation.
\newblock {\em ACM Transactions on Graphics (TOG)}, 39(6):1--15, 2020.

\bibitem{broxton2020immersive}
Michael Broxton, John Flynn, Ryan Overbeck, Daniel Erickson, Peter Hedman,
  Matthew Duvall, Jason Dourgarian, Jay Busch, Matt Whalen, and Paul Debevec.
\newblock Immersive light field video with a layered mesh representation.
\newblock {\em ACM Transactions on Graphics (TOG)}, 39(4):86--1, 2020.

\bibitem{buehler2001unstructured}
Chris Buehler, Michael Bosse, Leonard McMillan, Steven Gortler, and Michael
  Cohen.
\newblock Unstructured lumigraph rendering.
\newblock In {\em Proceedings of the 28th annual conference on Computer
  graphics and interactive techniques (CGIT)}, pages 425--432, 2001.

\bibitem{2022vr1}
G. Cai, K. Yan, Z. Dong, I. Gkioulekas, and S. Zhao.
\newblock Physics-based inverse rendering using combined implicit and explicit
  geometries.
\newblock In {\em Proceedings of the IEEE Conference on Computer Vision and
  Pattern Recognition (CVPR)}, 2022.

\bibitem{chai2000plenoptic}
Jin-Xiang Chai, Xin Tong, Shing-Chow Chan, and Heung-Yeung Shum.
\newblock Plenoptic sampling.
\newblock In {\em Proceedings of the 27th annual conference on Computer
  graphics and interactive techniques (CGIT)}, pages 307--318, 2000.

\bibitem{chaurasia2013depth}
Gaurav Chaurasia, Sylvain Duchene, Olga Sorkine-Hornung, and George Drettakis.
\newblock Depth synthesis and local warps for plausible image-based navigation.
\newblock {\em ACM Transactions on Graphics (TOG)}, 32(3):1--12, 2013.

\bibitem{chen2019learning}
Zhiqin Chen and Hao Zhang.
\newblock Learning implicit fields for generative shape modeling.
\newblock In {\em Proceedings of the IEEE Conference on Computer Vision and
  Pattern Recognition (CVPR)}, 2019.

\bibitem{choi2019extreme}
Inchang Choi, Orazio Gallo, Alejandro Troccoli, Min~H Kim, and Jan Kautz.
\newblock Extreme view synthesis.
\newblock In {\em Proceedings of the IEEE International Conference on Computer
  Vision (ICCV)}, 2019.

\bibitem{collet2015high}
Alvaro Collet, Ming Chuang, Pat Sweeney, Don Gillett, Dennis Evseev, David
  Calabrese, Hugues Hoppe, Adam Kirk, and Steve Sullivan.
\newblock High-quality streamable free-viewpoint video.
\newblock {\em ACM Transactions on Graphics (TOG)}, 34(4):1--13, 2015.

\bibitem{criminisi2007efficient}
Antonio Criminisi, Andrew Blake, Carsten Rother, Jamie Shotton, and Philip~HS
  Torr.
\newblock Efficient dense stereo with occlusions for new view-synthesis by
  four-state dynamic programming.
\newblock {\em International Journal of Computer Vision (IJCV)}, 71(1):89--110,
  2007.

\bibitem{2003Spacetime}
B Curless and S.~M Seitz.
\newblock Spacetime stereo: shape recovery for dynamic scenes.
\newblock In {\em Proceedings of the IEEE Conference on Computer Vision and
  Pattern Recognition (CVPR)}, 2003.

\bibitem{debevec1996modeling}
Paul~E Debevec, Camillo~J Taylor, and Jitendra Malik.
\newblock Modeling and rendering architecture from photographs: A hybrid
  geometry-and image-based approach.
\newblock In {\em Proceedings of the 23rd annual conference on Computer
  graphics and interactive techniques (CGIT)}, pages 11--20, 1996.

\bibitem{deng2022depth}
Kangle Deng, Andrew Liu, Jun-Yan Zhu, and Deva Ramanan.
\newblock Depth-supervised nerf: Fewer views and faster training for free.
\newblock In {\em Proceedings of the IEEE Conference on Computer Vision and
  Pattern Recognition (CVPR)}, 2022.

\bibitem{dosovitskiy2020image}
Alexey Dosovitskiy, Lucas Beyer, Alexander Kolesnikov, Dirk Weissenborn,
  Xiaohua Zhai, Thomas Unterthiner, Mostafa Dehghani, Matthias Minderer, Georg
  Heigold, Sylvain Gelly, et~al.
\newblock An image is worth 16x16 words: Transformers for image recognition at
  scale.
\newblock {\em arXiv preprint arXiv:2010.11929}, 2020.

\bibitem{2021NRF4D}
Yilun Du, Yinan Zhang, Hong~Xing Yu, Joshua~B. Tenenbaum, and Jiajun Wu.
\newblock Neural radiance flow for 4d view synthesis and video processing.
\newblock In {\em Proceedings of the IEEE International Conference on Computer
  Vision (ICCV)}, 2021.

\bibitem{2019DeepView}
J. Flynn, M. Broxton, P. Debevec, M. Duvall, G. Fyffe, R. Overbeck, N. Snavely,
  and R. Tucker.
\newblock Deepview: View synthesis with learned gradient descent.
\newblock In {\em Proceedings of the IEEE Conference on Computer Vision and
  Pattern Recognition (CVPR)}, 2019.

\bibitem{flynn2018deepstereo}
John Flynn, Keith Snavely, Ivan Neulander, and James Philbin.
\newblock Deepstereo: learning to predict new views from real world imagery,
  Mar.~13 2018.
\newblock US Patent 9,916,679.

\bibitem{2021Gao}
C. Gao, A. Saraf, J. Kopf, and J.~B. Huang.
\newblock Dynamic view synthesis from dynamic monocular video.
\newblock In {\em Proceedings of the IEEE International Conference on Computer
  Vision (ICCV)}, 2021.

\bibitem{gortler1996lumigraph}
Steven~J Gortler, Radek Grzeszczuk, Richard Szeliski, and Michael~F Cohen.
\newblock The lumigraph.
\newblock In {\em Proceedings of the 23rd annual conference on Computer
  graphics and interactive techniques (CGIT)}, pages 43--54, 1996.

\bibitem{hedman2017casual}
Peter Hedman, Suhib Alsisan, Richard Szeliski, and Johannes Kopf.
\newblock Casual 3d photography.
\newblock {\em ACM Transactions on Graphics (TOG)}, 36(6):1--15, 2017.

\bibitem{hedman2018deep}
Peter Hedman, Julien Philip, True Price, Jan-Michael Frahm, George Drettakis,
  and Gabriel Brostow.
\newblock Deep blending for free-viewpoint image-based rendering.
\newblock {\em ACM Transactions on Graphics (TOG)}, 37(6):1--15, 2018.

\bibitem{2017Single}
Philipp Henzler, Volker Rassche, Timo Ropinski, and Tobias Ritschel.
\newblock Single-image tomography: 3d volumes from 2d x-rays.
\newblock {\em Computer Graphics Forum (CGF)}, 37(2), 2017.

\bibitem{huang2018deep}
Zeng Huang, Tianye Li, Weikai Chen, Yajie Zhao, Jun Xing, Chloe LeGendre,
  Linjie Luo, Chongyang Ma, and Hao Li.
\newblock Deep volumetric video from very sparse multi-view performance
  capture.
\newblock In {\em Proceedings of the European Conference on Computer Vision
  (ECCV)}, 2018.

\bibitem{ivanov2013theory}
Valentin~K Ivanov, Vladimir~V Vasin, and Vitalii~P Tanana.
\newblock {\em Theory of Linear Ill-Posed Problems and its Applications},
  volume~36.
\newblock Walter de Gruyter, 2013.

\bibitem{2022NeuMan}
W. Jiang, K.~M. Yi, G. Samei, O. Tuzel, and A. Ranjan.
\newblock Neuman: Neural human radiance field from a single video.
\newblock In {\em Proceedings of the European Conference on Computer Vision
  (ECCV)}, 2022.

\bibitem{kanade1997virtualized}
Takeo Kanade, Peter Rander, and PJ Narayanan.
\newblock Virtualized reality: Constructing virtual worlds from real scenes.
\newblock {\em IEEE Transactions on Multimedia (MM)}, 4(1):34--47, 1997.

\bibitem{2017Learning}
Abhishek Kar, Christian Hne, and Jitendra Malik.
\newblock Learning a multi-view stereo machine.
\newblock In {\em Advances in Neural Information Processing Systems (NeurIPS)},
  2017.

\bibitem{kingma2014adam}
Diederik~P Kingma and Jimmy Ba.
\newblock Adam: A method for stochastic optimization.
\newblock {\em arXiv preprint arXiv:1412.6980}, 2014.

\bibitem{kingma2013auto}
Diederik~P Kingma and Max Welling.
\newblock Auto-encoding variational bayes.
\newblock {\em arXiv preprint arXiv:1312.6114}, 2013.

\bibitem{2022Panoptic}
A. Kundu, K. Genova, X. Yin, A. Fathi, C. Pantofaru, L. Guibas, A.
  Tagliasacchi, F. Dellaert, and T. Funkhouser.
\newblock Panoptic neural fields: A semantic object-aware neural scene
  representation.
\newblock In {\em Proceedings of the IEEE Conference on Computer Vision and
  Pattern Recognition (CVPR)}, 2022.

\bibitem{2020Blind}
Chenyang Lei, Yazhou Xing, and Qifeng Chen.
\newblock Blind video temporal consistency via deep video prior.
\newblock In {\em Advances in Neural Information Processing Systems (NeurIPS)},
  2020.

\bibitem{levoy1996light}
Marc Levoy and Pat Hanrahan.
\newblock Light field rendering.
\newblock In {\em Proceedings of the 23rd annual conference on Computer
  graphics and interactive techniques (CGIT)}, pages 31--42, 1996.

\bibitem{li2012temporally}
Hao Li, Linjie Luo, Daniel Vlasic, Pieter Peers, Jovan Popovi{\'c}, Mark Pauly,
  and Szymon Rusinkiewicz.
\newblock Temporally coherent completion of dynamic shapes.
\newblock {\em ACM Transactions on Graphics (TOG)}, 31(1):1--11, 2012.

\bibitem{2022Multiview}
Tianye Li, Mira Slavcheva, Michael Zollhoefer, Simon Green, Christoph Lassner,
  Changil Kim, Tanner Schmidt, Steven Lovegrove, Michael Goesele, Richard
  Newcombe, and Zhaoyang Lv.
\newblock Neural {3D} video synthesis from multi-view video.
\newblock In {\em Proceedings of the IEEE Conference on Computer Vision and
  Pattern Recognition (CVPR)}, 2022.

\bibitem{2021NSFF}
Zhengqi Li, Simon Niklaus, Noah Snavely, and Oliver Wang.
\newblock Neural scene flow fields for space-time view synthesis of dynamic
  scenes.
\newblock In {\em Proceedings of the IEEE Conference on Computer Vision and
  Pattern Recognition (CVPR)}, 2021.

\bibitem{liu2021swin}
Ze Liu, Yutong Lin, Yue Cao, Han Hu, Yixuan Wei, Zheng Zhang, Stephen Lin, and
  Baining Guo.
\newblock Swin transformer: Hierarchical vision transformer using shifted
  windows.
\newblock In {\em Proceedings of the IEEE International Conference on Computer
  Vision (ICCV)}, 2021.

\bibitem{2020Consistent}
Xuan Luo, Jia{-}Bin Huang, Richard Szeliski, Kevin Matzen, and Johannes Kopf.
\newblock Consistent video depth estimation.
\newblock In {\em ACM Transactions on Graphics (Proceedings of ACM SIGGRAPH)},
  volume~39. ACM, 2020.

\bibitem{2021Nerfw}
Ricardo Martin-Brualla, Noha Radwan, Mehdi S.~M. Sajjadi, Jonathan~T. Barron,
  Alexey Dosovitskiy, and Daniel Duckworth.
\newblock Nerf in the wild: Neural radiance fields for unconstrained photo
  collections.
\newblock In {\em Proceedings of the IEEE Conference on Computer Vision and
  Pattern Recognition (CVPR)}, 2021.

\bibitem{mescheder2019occupancy}
Lars Mescheder, Michael Oechsle, Michael Niemeyer, Sebastian Nowozin, and
  Andreas Geiger.
\newblock Occupancy networks: Learning 3d reconstruction in function space.
\newblock In {\em Proceedings of the IEEE Conference on Computer Vision and
  Pattern Recognition (CVPR)}, 2019.

\bibitem{michalkiewicz2019deep}
Mateusz Michalkiewicz, Jhony~K Pontes, Dominic Jack, Mahsa Baktashmotlagh, and
  Anders Eriksson.
\newblock Deep level sets: Implicit surface representations for 3d shape
  inference.
\newblock {\em arXiv preprint arXiv:1901.06802}, 2019.

\bibitem{2022Dark}
B. Mildenhall, P. Hedman, R. Martin-Brualla, P. Srinivasan, and J.~T. Barron.
\newblock Nerf in the dark: High dynamic range view synthesis from noisy raw
  images.
\newblock In {\em Proceedings of the IEEE Conference on Computer Vision and
  Pattern Recognition (CVPR)}, 2022.

\bibitem{mildenhall2019local}
Ben Mildenhall, Pratul~P Srinivasan, Rodrigo Ortiz-Cayon, Nima~Khademi
  Kalantari, Ravi Ramamoorthi, Ren Ng, and Abhishek Kar.
\newblock Local light field fusion: Practical view synthesis with prescriptive
  sampling guidelines.
\newblock {\em ACM Transactions on Graphics (TOG)}, 38(4):1--14, 2019.

\bibitem{2020Nerf}
Ben Mildenhall, Pratul~P. Srinivasan, Matthew Tancik, Jonathan~T. Barron, Ravi
  Ramamoorthi, and Ren Ng.
\newblock Nerf: Representing scenes as neural radiance fields for view
  synthesis.
\newblock In {\em Proceedings of the European Conference on Computer Vision
  (ECCV)}, 2020.

\bibitem{2019Differentiable}
Michael Niemeyer, Lars Mescheder, Michael Oechsle, and Andreas Geiger.
\newblock Differentiable volumetric rendering: Learning implicit 3d
  representations without 3d supervision.
\newblock In {\em Proceedings of the IEEE Conference on Computer Vision and
  Pattern Recognition (CVPR)}, 2019.

\bibitem{niklaus20193d}
Simon Niklaus, Long Mai, Jimei Yang, and Feng Liu.
\newblock 3d ken burns effect from a single image.
\newblock {\em ACM Transactions on Graphics (TOG)}, 38(6):1--15, 2019.

\bibitem{orts2016holoportation}
Sergio Orts-Escolano, Christoph Rhemann, Sean Fanello, Wayne Chang, Adarsh
  Kowdle, Yury Degtyarev, David Kim, Philip~L Davidson, Sameh Khamis, Mingsong
  Dou, et~al.
\newblock Holoportation: Virtual 3d teleportation in real-time.
\newblock In {\em Proceedings of the 29th Annual Symposium on User Interface
  Software and Technology (UIST)}, pages 741--754, 2016.

\bibitem{oswald2014generalized}
Martin~Ralf Oswald, Jan St{\"u}hmer, and Daniel Cremers.
\newblock Generalized connectivity constraints for spatio-temporal 3d
  reconstruction.
\newblock In {\em Proceedings of the European Conference on Computer Vision
  (ECCV)}, 2014.

\bibitem{2018vr2}
Ryan~S. Overbeck, Daniel Erickson, Daniel Evangelakos, and Paul Debevec.
\newblock The making of welcome to light fields vr.
\newblock In {\em ACM Transactions on Graphics (Proceedings of ACM SIGGRAPH)},
  pages 1--2. ACM, 2018.

\bibitem{park2019deepsdf}
Jeong~Joon Park, Peter Florence, Julian Straub, Richard Newcombe, and Steven
  Lovegrove.
\newblock Deepsdf: Learning continuous signed distance functions for shape
  representation.
\newblock In {\em Proceedings of the IEEE Conference on Computer Vision and
  Pattern Recognition (CVPR)}, 2019.

\bibitem{2021Nerfies}
Keunhong Park, Utkarsh Sinha, Jonathan~T. Barron, Sofien Bouaziz, Dan~B
  Goldman, Steven~M. Seitz, and Ricardo Martin-Brualla.
\newblock Nerfies: Deformable neural radiance fields.
\newblock In {\em Proceedings of the IEEE International Conference on Computer
  Vision (ICCV)}, 2021.

\bibitem{park2021hypernerf}
Keunhong Park, Utkarsh Sinha, Peter Hedman, Jonathan~T Barron, Sofien Bouaziz,
  Dan~B Goldman, Ricardo Martin-Brualla, and Steven~M Seitz.
\newblock Hypernerf: A higher-dimensional representation for topologically
  varying neural radiance fields.
\newblock {\em arXiv preprint arXiv:2106.13228}, 2021.

\bibitem{penner2017soft}
Eric Penner and Li Zhang.
\newblock Soft 3d reconstruction for view synthesis.
\newblock {\em ACM Transactions on Graphics (TOG)}, 36(6):1--11, 2017.

\bibitem{2021DNeRF}
Albert Pumarola, Enric Corona, Gerard Pons-Moll, and Francesc Moreno-Noguer.
\newblock D-nerf: Neural radiance fields for dynamic scenes.
\newblock In {\em Proceedings of the IEEE Conference on Computer Vision and
  Pattern Recognition (CVPR)}, 2021.

\bibitem{ranftl2020towards}
Ren{\'e} Ranftl, Katrin Lasinger, David Hafner, Konrad Schindler, and Vladlen
  Koltun.
\newblock Towards robust monocular depth estimation: Mixing datasets for
  zero-shot cross-dataset transfer.
\newblock {\em IEEE Transactions on Pattern Analysis and Machine Intelligence
  (TPAMI)}, 2020.

\bibitem{2022Urban}
K. Rematas, A. Liu, P.~P. Srinivasan, J.~T. Barron, A. Tagliasacchi, T.
  Funkhouser, and V. Ferrari.
\newblock Urban radiance fields.
\newblock In {\em Proceedings of the IEEE Conference on Computer Vision and
  Pattern Recognition (CVPR)}, 2022.

\bibitem{riegler2020free}
Gernot Riegler and Vladlen Koltun.
\newblock Free view synthesis.
\newblock In {\em Proceedings of the European Conference on Computer Vision
  (ECCV)}, 2020.

\bibitem{schonberger2016structure}
Johannes~L Schonberger and Jan-Michael Frahm.
\newblock Structure-from-motion revisited.
\newblock In {\em Proceedings of the IEEE Conference on Computer Vision and
  Pattern Recognition (CVPR)}, 2016.

\bibitem{shi2014light}
Lixin Shi, Haitham Hassanieh, Abe Davis, Dina Katabi, and Fredo Durand.
\newblock Light field reconstruction using sparsity in the continuous fourier
  domain.
\newblock {\em ACM Transactions on Graphics (TOG)}, 34(1):1--13, 2014.

\bibitem{snavely2006photo}
Noah Snavely, Steven~M Seitz, and Richard Szeliski.
\newblock Photo tourism: exploring photo collections in 3d.
\newblock In {\em ACM Transactions on Graphics (Proceedings of ACM SIGGRAPH)},
  pages 835--846. ACM, 2006.

\bibitem{2019Pushing}
Pratul~P. Srinivasan, Richard Tucker, Jonathan~T. Barron, Ravi Ramamoorthi, Ren
  Ng, and Noah Snavely.
\newblock Pushing the boundaries of view extrapolation with multiplane images.
\newblock In {\em Proceedings of the IEEE Conference on Computer Vision and
  Pattern Recognition (CVPR)}, 2019.

\bibitem{srinivasan2019pushing}
Pratul~P Srinivasan, Richard Tucker, Jonathan~T Barron, Ravi Ramamoorthi, Ren
  Ng, and Noah Snavely.
\newblock Pushing the boundaries of view extrapolation with multiplane images.
\newblock In {\em Proceedings of the IEEE Conference on Computer Vision and
  Pattern Recognition (CVPR)}, 2019.

\bibitem{2022Block}
M. Tancik, V. Casser, X. Yan, S. Pradhan, B. Mildenhall, P.~P. Srinivasan,
  J.~T. Barron, and H. Kretzschmar.
\newblock Block-nerf: Scalable large scene neural view synthesis.
\newblock In {\em Proceedings of the IEEE Conference on Computer Vision and
  Pattern Recognition (CVPR)}, 2022.

\bibitem{tancik2020fourier}
Matthew Tancik, Pratul Srinivasan, Ben Mildenhall, Sara Fridovich-Keil, Nithin
  Raghavan, Utkarsh Singhal, Ravi Ramamoorthi, Jonathan Barron, and Ren Ng.
\newblock Fourier features let networks learn high frequency functions in low
  dimensional domains.
\newblock In {\em Advances in Neural Information Processing Systems (NeurIPS)},
  2020.

\bibitem{teed2020raft}
Zachary Teed and Jia Deng.
\newblock Raft: Recurrent all-pairs field transforms for optical flow.
\newblock In {\em Proceedings of the European Conference on Computer Vision
  (ECCV)}, 2020.

\bibitem{tewari2020state}
Ayush Tewari, Ohad Fried, Justus Thies, Vincent Sitzmann, Stephen Lombardi,
  Kalyan Sunkavalli, Ricardo Martin-Brualla, Tomas Simon, Jason Saragih,
  Matthias Nie{\ss}ner, et~al.
\newblock State of the art on neural rendering.
\newblock In {\em Computer Graphics Forum (CGF)}, volume~39, pages 701--727.
  Wiley Online Library, 2020.

\bibitem{2021NonRigid}
Edgar Tretschk, Ayush Tewari, Vladislav Golyanik, Michael Zollhfer, Christoph
  Lassner, and Christian Theobalt.
\newblock Non-rigid neural radiance fields: Reconstruction and novel view
  synthesis of a dynamic scene from monocular video.
\newblock In {\em Proceedings of the IEEE International Conference on Computer
  Vision (ICCV)}, 2021.

\bibitem{trevithick2020grf}
Alex Trevithick and Bo Yang.
\newblock Grf: Learning a general radiance field for 3d representation and
  rendering.
\newblock {\em arXiv preprint arXiv:2010.04595}, 2020.

\bibitem{tucker2020single}
Richard Tucker and Noah Snavely.
\newblock Single-view view synthesis with multiplane images.
\newblock In {\em Proceedings of the IEEE Conference on Computer Vision and
  Pattern Recognition (CVPR)}, 2020.

\bibitem{2022Mega}
H. Turki, D. Ramanan, and M. Satyanarayanan.
\newblock Mega-nerf: Scalable construction of large-scale nerfs for virtual
  fly-throughs.
\newblock In {\em Proceedings of the IEEE Conference on Computer Vision and
  Pattern Recognition (CVPR)}, 2022.

\bibitem{vaswani2017attention}
Ashish Vaswani, Noam Shazeer, Niki Parmar, Jakob Uszkoreit, Llion Jones,
  Aidan~N Gomez, {\L}ukasz Kaiser, and Illia Polosukhin.
\newblock Attention is all you need.
\newblock In {\em Advances in Neural Information Processing Systems (NeurIPS)},
  2017.

\bibitem{2022Fourier}
L. Wang, J. Zhang, X. Liu, F. Zhao, Y. Zhang, Y. Zhang, M. Wu, L. Xu, and J.
  Yu.
\newblock Fourier plenoctrees for dynamic radiance field rendering in
  real-time.
\newblock In {\em Proceedings of the IEEE Conference on Computer Vision and
  Pattern Recognition (CVPR)}, 2022.

\bibitem{wang2004image}
Zhou Wang, Alan~C Bovik, Hamid~R Sheikh, and Eero~P Simoncelli.
\newblock Image quality assessment: from error visibility to structural
  similarity.
\newblock {\em IEEE Transactions on Image Processing (TIP)}, 13(4):600--612,
  2004.

\bibitem{Argoverse2}
Benjamin Wilson, William Qi, Tanmay Agarwal, John Lambert, Jagjeet Singh,
  Siddhesh Khandelwal, Bowen Pan, Ratnesh Kumar, Andrew Hartnett,
  Jhony~Kaesemodel Pontes, Deva Ramanan, Peter Carr, and James Hays.
\newblock Argoverse 2: Next generation datasets for self-driving perception and
  forecasting.
\newblock In {\em Proceedings of the Neural Information Processing Systems
  Track on Datasets and Benchmarks (NeurIPS Datasets and Benchmarks 2021)},
  2021.

\bibitem{2021Space}
Wenqi Xian, Jia~Bin Huang, Johannes Kopf, and Changil Kim.
\newblock Space-time neural irradiance fields for free-viewpoint video.
\newblock In {\em Proceedings of the IEEE Conference on Computer Vision and
  Pattern Recognition (CVPR)}, 2021.

\bibitem{2022Bungee}
Yuanbo Xiangli, Linning Xu, Xingang Pan, Nanxuan Zhao, Anyi Rao, Christian
  Theobalt, Bo Dai, and Dahua Lin.
\newblock Bungeenerf: Progressive neural radiance field for extreme multi-scale
  scene rendering.
\newblock In {\em Proceedings of the European Conference on Computer Vision
  (ECCV)}, 2022.

\bibitem{xu2019deep}
Zexiang Xu, Sai Bi, Kalyan Sunkavalli, Sunil Hadap, Hao Su, and Ravi
  Ramamoorthi.
\newblock Deep view synthesis from sparse photometric images.
\newblock {\em ACM Transactions on Graphics (TOG)}, 38(4):1--13, 2019.

\bibitem{2020Yoon}
Jae~Shin Yoon, Kihwan Kim, Orazio Gallo, Hyun~Soo Park, and Jan Kautz.
\newblock Novel view synthesis of dynamic scenes with globally coherent depths
  from a monocular camera.
\newblock In {\em Proceedings of the IEEE Conference on Computer Vision and
  Pattern Recognition (CVPR)}, 2020.

\bibitem{yuan2021star}
Wentao Yuan, Zhaoyang Lv, Tanner Schmidt, and Steven Lovegrove.
\newblock Star: Self-supervised tracking and reconstruction of rigid objects in
  motion with neural rendering.
\newblock In {\em Proceedings of the IEEE Conference on Computer Vision and
  Pattern Recognition (CVPR)}, 2021.

\bibitem{2021Nerf++}
Kai Zhang, Gernot Riegler, Noah Snavely, and Vladlen Koltun.
\newblock Nerf++: Analyzing and improving neural radiance fields.
\newblock {\em arXiv:2010.07492}, 2021.

\bibitem{zhang2018perceptual}
Richard Zhang, Phillip Isola, Alexei~A Efros, Eli Shechtman, and Oliver Wang.
\newblock The unreasonable effectiveness of deep features as a perceptual
  metric.
\newblock In {\em Proceedings of the IEEE Conference on Computer Vision and
  Pattern Recognition (CVPR)}, 2018.

\bibitem{2018Stereo}
Tinghui Zhou, Richard Tucker, John Flynn, Graham Fyffe, and Noah Snavely.
\newblock Stereo magnification: learning view synthesis using multiplane
  images.
\newblock {\em ACM Transactions on Graphics (Proceedings of ACM SIGGRAPH)}, 4,
  2018.

\bibitem{zhou2018stereo}
Tinghui Zhou, Richard Tucker, John Flynn, Graham Fyffe, and Noah Snavely.
\newblock Stereo magnification: Learning view synthesis using multiplane
  images.
\newblock {\em arXiv preprint arXiv:1805.09817}, 2018.

\bibitem{zitnick2004high}
C~Lawrence Zitnick, Sing~Bing Kang, Matthew Uyttendaele, Simon Winder, and
  Richard Szeliski.
\newblock High-quality video view interpolation using a layered representation.
\newblock {\em ACM Transactions on Graphics (TOG)}, 23(3):600--608, 2004.

\end{thebibliography}
}

\end{document}